\documentclass{bmvc2k}
\usepackage{url}
\usepackage{graphicx}
\usepackage{booktabs}
\usepackage[table]{xcolor}
\usepackage{multirow}
\usepackage[ruled,vlined]{algorithm2e}
\usepackage{amssymb} 
\usepackage{mathtools}
\usepackage{minted}
\usepackage{wrapfig}
\usepackage{makecell}
\usepackage{enumitem}

\title{Relax Forcing: Relaxed KV-Memory for Consistent Long Video Generation} 

\addauthor{Zengqun Zhao}{zengqun.zhao@qmul.ac.uk}{1}
\addauthor{Yanzuo Lu}{oliveryanzuolu@gmail.com}{2}
\addauthor{Ziquan Liu}{ziquan.liu@qmul.ac.uk}{1}
\addauthor{Jifei Song}{jifeisong@huawei.com}{3}
\addauthor{Jiankang Deng}{j.deng16@imperial.ac.uk}{2}
\addauthor{Ioannis Patras}{i.patras@qmul.ac.uk}{1}

\addinstitution{Queen Mary University of London}
\addinstitution{Imperial College London}
\addinstitution{Huawei R\&D UK}

\runninghead{Zhao et al.}{Relax Forcing}

\begin{document}
\maketitle

\vspace{-1.5em}
\begin{abstract}
Autoregressive video diffusion has recently emerged as a promising paradigm for long-video generation, enabling causal synthesis beyond the temporal limits of bidirectional models. Existing forcing-based training strategies reduce exposure bias by conditioning models on their own predictions during rollout, yet minute-scale generation remains challenging due to progressive temporal degradation and constrained motion evolution. In this work, we study the role of temporal KV memory during long-horizon autoregressive inference. Our analysis shows that simply retaining more historical frames does not consistently improve generation quality; instead, both the quantity and temporal placement of memory strongly affect motion dynamics. These findings suggest that temporal memory should be treated as structured context rather than a homogeneous chronological buffer. Motivated by this observation, we introduce Relax Forcing, a training-free memory mechanism for autoregressive video diffusion. Relax Forcing decomposes temporal context into three functional components: Sink frames that provide global stability, Tail frames that preserve short-term continuity, and dynamically selected History frames that supply mid-range motion structure. History frames are selected using a relaxation-based criterion that encourages alignment with global anchors while suppressing redundancy with recent context. This role-aware sparse memory design mitigates error accumulation during long-horizon rollout while preserving motion evolution and reducing attention overhead. Experiments on VBench-Long show that Relax Forcing improves long-video generation, achieving stronger motion dynamics and higher overall scores than existing autoregressive baselines. These results indicate that structured temporal memory is an effective and complementary direction for scalable long-video generation. \url{https://zengqunzhao.github.io/Relax-Forcing}
\end{abstract}

\vspace{-1.8em}
\section{Introduction}
\vspace{-0.7em}
\label{sec:intro}

\begin{figure}[!t]
    \centering 
    \includegraphics[width=0.99\textwidth]{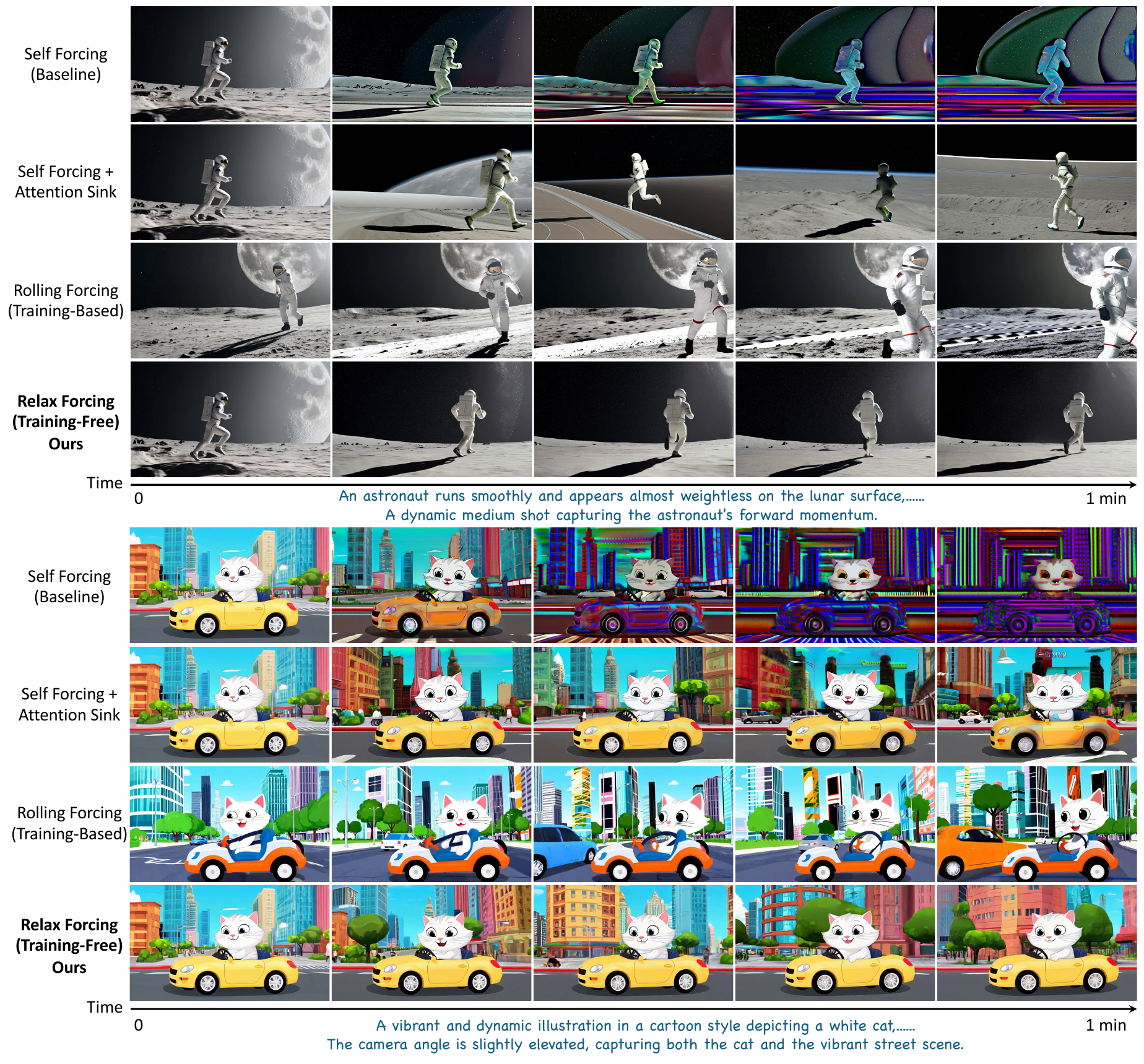}
    \vspace{-1em}
    \caption{Long-horizon autoregressive video generation over one minute. Baseline Self Forcing suffers from severe temporal drift and visual degradation as errors accumulate. Adding an attention sink stabilises early frames but limits motion diversity. Rolling Forcing improves short-term coherence but still exhibits inconsistent patterns. In contrast, Relax Forcing dynamically selects informative history while suppressing redundancy, achieving stable identity preservation and sustained scene evolution.}
    \label{fig1}
    \vspace{-1.5em}
\end{figure}
Recent advances in video diffusion models have enabled high-fidelity short video synthesis~\cite{blattmann2023stable,yang2024cogvideox,kong2024hunyuanvideo,wang2025wan,zhao2026latsearch}. However, many interactive~\cite{yang2025longlive,shin2025motionstream,yu2025context} and streaming~\cite{bruce2024genie,ye2025yan,hong2025relic} applications require videos to be generated causally over minute-scale horizons, where future frames are unavailable and the generation length may exceed the training sequence length. Autoregressive (AR) video diffusion provides a practical solution by synthesising frames sequentially and reusing past context through key–value (KV) caching~\cite{chen2024diffusion,sun2025ar,yin2025slow,huang2025self,liu2025rolling}. To reduce the resulting training–inference mismatch, rollout-based training strategies such as Self Forcing~\cite{huang2025self} and its extensions~\cite{cui2025self,lu2025reward,zhang2026reward,po2025bagger,zhu2026causal,chen2026context} train models to condition on their own generated frames. At inference time, these models are typically rolled out with a sliding-window context: newly generated frames are appended to memory, while older frames are discarded once the window is saturated. This mechanism enables generation beyond the training sequence length, but also makes long-horizon quality depend critically on how historical context is retained and reused.

While rollout-based training substantially improves robustness, extending AR diffusion to minute-scale synthesis remains challenging, a problem commonly referred to as long-video extrapolation~\cite{zhao2025riflex,zhao2025ultravico}. Even with reduced training–inference mismatch, autoregressive models repeatedly condition on their own generated frames, allowing residual errors and temporal biases to accumulate over time. As a result, long-horizon generation often exhibits gradual drift, visual degradation, or overly constrained motion dynamics. Recent work has therefore begun to shift attention from training strategies to inference-time memory management~\cite{dalal2025one,zhang2025frame,yesiltepe2025infinity,hong2025relic,yu2025videossm,yi2025deep,ji2025memflow,zhang2025pretraining,zhu2025memorize,yu2025context,cui2026lol,chen2026past,chen2026context}. For example, Deep Forcing~\cite{yi2025deep} compresses candidate memories based on query-averaged attention scores, while Reward Forcing~\cite{lu2025reward} preserves long-range historical information through an exponential moving average mechanism. Other methods summarise long video histories into compact contexts through memory compression or adaptive memory mechanisms~\cite{zhang2025pretraining,zhu2025memorize}. Despite these advances, existing approaches primarily treat memory as a capacity, compression, or update problem. Less attention has been paid to the functional roles of different temporal regions: which historical frames stabilise appearance, which preserve local continuity, and which support future motion evolution.

\begin{figure}[t]
\centering
\includegraphics[width=0.99\textwidth]{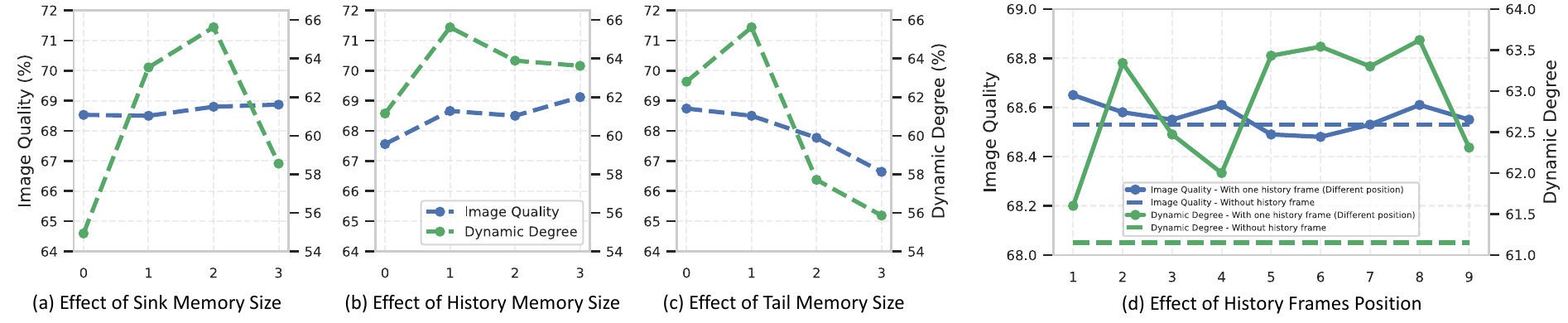}
\vspace{-1.1em}
\caption{Analysis of temporal memory in long-video extrapolation. (a–c): Varying the number of Sink, History, and Tail frames shows that increasing conditioning memory size does not consistently improve generation quality and may constrain motion dynamics. (d): Under a fixed memory budget, selecting history frames from different temporal positions leads to noticeable variations in motion dynamics while leaving image quality largely unchanged.}
\label{fig2}
\vspace{-1.7em}
\end{figure}

To better understand this issue, we analyse how temporal memory configurations affect long-video extrapolation. Our results show that dense memory is not uniformly beneficial; instead, its effectiveness depends on both the amount of retained context and its temporal placement. As shown in Fig.~\ref{fig2}(a–c), increasing the number of conditioning frames from Sink, mid-range History, or Tail does not consistently improve long-horizon generation. Beyond a certain point, additional memory introduces redundant temporal cues that weaken motion dynamics without yielding meaningful gains in visual quality. Moreover, Fig.~\ref{fig2}(d) shows that, under a fixed memory budget, sampling history frames from different temporal positions leads to different motion dynamics while leaving image quality largely unchanged. These observations suggest that historical frames are not interchangeable. Rather than acting as a homogeneous chronological buffer, temporal memory contains distinct functional regions: early frames can serve as global anchors, recent frames preserve short-term continuity, and mid-range frames provide structural cues for motion evolution.

\begin{figure}[t]
\vspace{-0.3em}
\centering
\includegraphics[width=0.98\textwidth]{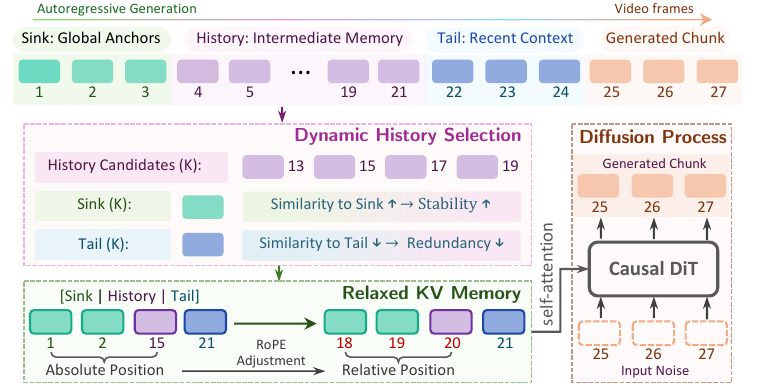}
\caption{Overview of Relaxed KV Memory. Instead of retaining dense chronological history, temporal memory is decomposed into three functional components: \emph{Sink} for global anchors, \emph{History} for intermediate motion structure, and \emph{Tail} for recent continuity. During generation, candidate historical frames are dynamically selected to remain aligned with Sink while avoiding redundancy with Tail. The selected memory is then integrated through a relaxed KV formulation with adjusted relative positional encoding, enabling the model to leverage non-contiguous temporal context while preserving long-range consistency during autoregressive rollout.}
\label{fig4}
\vspace{-1.5em}
\end{figure}

Motivated by this role-specific view, we propose Relaxed KV Memory, a structured sparse memory mechanism for AR diffusion, as shown in Fig.~\ref{fig4}. We refer to this training-free inference framework as Relax Forcing. Instead of attending to a dense chronological history, our method decomposes temporal context into three components: \emph{Sink} for long-term stability, \emph{Tail} for short-term continuity, and dynamically selected \emph{History} for mid-range motion structure. The History frames are selected by a relaxation-based scoring mechanism that favours alignment with global Sink anchors while penalising redundancy with recent Tail context. This criterion differs from generic cache compression or attention-based memory selection: it explicitly selects historical frames according to their temporal role in balancing stability and motion evolution. By retaining only complementary historical cues, Relaxed KV Memory reduces over-reliance on recent context, mitigates error accumulation during extrapolation, and reduces the effective attention length during long-horizon rollout.

In summary, the main contributions of this work are as follows:
\vspace{-0.9em}
\begin{itemize}
\item We conduct a systematic analysis of how temporal memory affects long-video extrapolation in AR diffusion. Our study shows that increasing memory size alone is insufficient and that the temporal placement of retained frames plays a critical role in long-horizon motion evolution.
\vspace{-0.9em}
\item We propose Relaxed KV Memory, a training-free, role-aware sparse memory mechanism that decomposes temporal context into Sink, History, and Tail components. The proposed relaxation score selects mid-range history that remains aligned with global anchors while avoiding redundancy with recent frames.
\vspace{-0.9em}
\item Experiments on VBench-Long show that our approach improves long-video generation, achieving a 66.8\% relative improvement in Dynamic Degree and a 5.7\% relative gain in the overall score over the Self Forcing baseline, while reducing effective attention length and improving inference efficiency.
\end{itemize}

\vspace{-2.5em}
\section{Related Work}
\vspace{-0.5em}
\label{sec:related}
\noindent \textbf{From Bidirectional to Autoregressive Video Diffusion.}
Recent advances in video generation have been largely driven by diffusion models that denoise all frames simultaneously using bidirectional attention, enabling strong visual fidelity but requiring access to the full sequence during generation \cite{blattmann2023stable,yang2024cogvideox,kong2024hunyuanvideo,wang2025wan}. This design limits scalability to real-time or long-horizon scenarios where future frames are unavailable. To address this, a growing body of work has explored autoregressive (AR) formulations of video diffusion, enabling sequential generation that aligns with the causal structure of temporal data \cite{sun2025ar,yin2025slow,li2026skyreels}. Early attempts reformulate diffusion training using teacher-forcing-style next-frame prediction \cite{gao2024ca2,zhou2025taming} or noise-independent training schemes such as diffusion forcing \cite{chen2024diffusion} and FAR \cite{gu2025long}. Other methods also introduce queue-based conditioning and asynchronous generation strategies to extend temporal rollout, including FIFO-style history replacement \cite{kim2024fifo}, asynchronous AR diffusion \cite{sun2025ar}, and masked autoregressive training \cite{zhou2025taming}. History-aware conditioning \cite{song2025history}, fast AR conversion from bidirectional models \cite{yin2025slow}, and industrial-scale AR systems \cite{li2026skyreels,teng2025magi} further demonstrate the feasibility of long-form sequential generation. However, these approaches inherently introduce a training–inference mismatch: models are typically trained with clean or distribution-smoothed historical context but must condition on their own imperfect predictions during autoregressive rollout. This mismatch leads to error accumulation over time, commonly referred to as exposure bias, motivating the development of rollout-aware training paradigms.

\noindent \textbf{Mitigating Exposure Bias in AR Video Diffusion.}
To bridge the training-inference gap, recent work increasingly moves beyond ground-truth conditioning toward training regimes that explicitly model autoregressive dynamics. Self Forcing~\cite{huang2025self} unrolls generation during training and supervises the model on sequences conditioned on its own predictions, significantly reducing exposure bias. Self-Forcing++~\cite{cui2025self} extends this idea to longer horizons, enabling minute-scale video synthesis with improved temporal stability. Other methods explore dynamic context updates through rolling conditioning strategies~\cite{liu2025rolling} or error recycling mechanisms for infinite-length generation~\cite{li2025stable}. Backward aggregation mitigates long-term degradation by incorporating historical corrections~\cite{po2025bagger}, while reward-guided methods introduce external supervision during rollout to steer generation toward higher-quality trajectories~\cite{lu2025reward,zhang2026reward}. End-to-end AR optimisation via self-resampling further aligns training with inference-time dynamics~\cite{guo2025end}. Despite these advances, exposure-bias mitigation alone does not fully resolve long-horizon degradation. As generation horizons extend, stability and efficiency increasingly depend on how historical information is stored, prioritised, and reused during self-attention.

\noindent \textbf{Memory Management for Long Video Extrapolation.}
When AR diffusion is rolled out beyond its training horizon, temporal degradation becomes tightly coupled with how historical context is stored and reused. Existing approaches mainly address this issue through memory compression, cache selection, or adaptive memory updating. Memory compression methods reduce attention overhead by summarising long histories into compact representations, reorganising frame context, or compressing KV caches~\cite{zhang2025frame,zhang2025pretraining,zhu2025memorize}. Other methods introduce adaptive memory mechanisms or alternative state representations to maintain long-range consistency under constrained attention budgets~\cite{yu2025videossm,ji2025memflow,hong2025relic}. More recently, inference-time cache policies select or compress historical information according to attention scores, salience estimates, or long-context anchors~\cite{yi2025deep,chen2026past,chen2026context}. These methods improve scalability and reduce redundant computation, but they primarily ask which tokens or frames should be retained under a memory budget.
Our work studies a complementary question: the functional roles different temporal regions play during long-horizon rollout. We show that early, middle, and recent frames affect generation differently: early frames provide global anchors, recent frames preserve short-term continuity, and mid-range history supports motion evolution. Based on this observation, Relaxed KV Memory selects sparse temporal context according to role-specific criteria, favouring history frames that remain aligned with global anchors while avoiding redundancy with recent context. This differs from generic KV compression or attention-score-based selection, and provides a training-free mechanism that complements forcing-based training strategies.

\vspace{-1.5em}
\section{Methods}
\vspace{-0.5em}
\label{sec:method}
\subsection{Preliminaries}
\vspace{-0.5em}
\noindent \textbf{Autoregressive Video Diffusion.}~
Given a video sequence $\{x^1,...,x^N\}$, autoregressive (AR) video diffusion models the joint distribution as $p_\theta(x^{1:N}) = \prod_{i=1}^{N} p_\theta(x^i \mid c^{<i})$, where each conditional distribution is parameterised by a diffusion process that iteratively denoises a noisy latent $z_t^i$ into the target frame $x^i$ conditioned on historical context $c^{<i}$. 
The fundamental distinction among AR training paradigms lies in the formulation of this context. 
Under Teacher Forcing~\cite{gao2024ca2}, the model conditions on pristine ground-truth history $c^{<i}=x^{<i}\sim P_{data}$, enabling parallelised training but inducing severe exposure bias, as inference requires conditioning on accumulated model predictions $\hat{x}^{<i}\sim P_\theta$ rather than clean observations. 
Diffusion Forcing~\cite{chen2024diffusion} partially mitigates this mismatch by injecting continuous noise into the ground-truth context, defining $c^{<i}=z_t^{<i}\sim q(z_t^{<i}|x^{<i})$, yet isotropic Gaussian perturbations fail to replicate the structured sequential artifacts produced during actual autoregressive rollout. 
Self Forcing~\cite{huang2025self} achieves strict training-inference alignment by directly defining $c^{<i}=\hat{x}^{<i}\sim P_\theta$, unrolling generation during training such that the denoising objective for each target frame $x^i$ is optimised conditioned on frames sampled through the model's own reverse diffusion trajectory, forcing the network to directly observe and correct the structured artifacts it encounters at inference.

\noindent \textbf{Long Video Rollout and Extrapolation.}~
To synthesise sequences exceeding the fixed temporal capacity of the bidirectional teacher window, Self Forcing employs a chunk-wise sliding window strategy. 
Given a maximum window size $L$ partitioned into chunks of length $U$, the model progressively accumulates generated chunks within a window, expanding $c^{<i}=\hat{x}^{<i}\sim P_\theta$ until the capacity limit $L$ is reached. 
Upon saturation, the subsequent window is initialised by truncating the global context and conditioning solely on the final chunk of the preceding window as an overlapping temporal anchor $\hat{x}^{L-U:L}\sim P_\theta$, from which progressive chunk accumulation resumes. 
However, this rigid reliance on a fixed previously generated anchor forces residual errors and temporal biases to compound across successive window transitions, rendering long-horizon extrapolation susceptible to gradual semantic drift and overly constrained motion dynamics. This highlights the fundamental limitations of static memory conditioning.

\noindent \textbf{Rotary Position Embedding.}~
To manage positional information during sliding window extrapolation, Self Forcing employs a strictly localised RoPE~\cite{su2024roformer} strategy that deliberately discards global temporal progression. 
Upon each window transition, the overlapping anchor $\hat{x}^{L-U:L}$ has its positional indices forcibly reset to $[0, U-1]$, with newly synthesised frames assigned local indices $[U, L-1]$, rendering the model entirely agnostic to the total number of frames generated across prior rollouts. 
This localised coordinate mapping coincides with periodic KV cache overwriting at each window boundary, compelling the model to treat every extrapolation step as an independent short-horizon task. 
However, this rigid consecutive index reassignment does not preserve the true relative temporal distances between frames, artificially compressing temporal gaps and forcing historically distant frames to appear adjacent to recent ones. 
Consequently, this inflexible RoPE mapping structurally prevents the model from leveraging long-range historical context, severely limiting its capacity to preserve global visual coherence and accurate motion dynamics over extended durations.

\begin{figure}[t]
    \vspace{-0.5em}
    \centering 
    \includegraphics[width=0.99\textwidth]{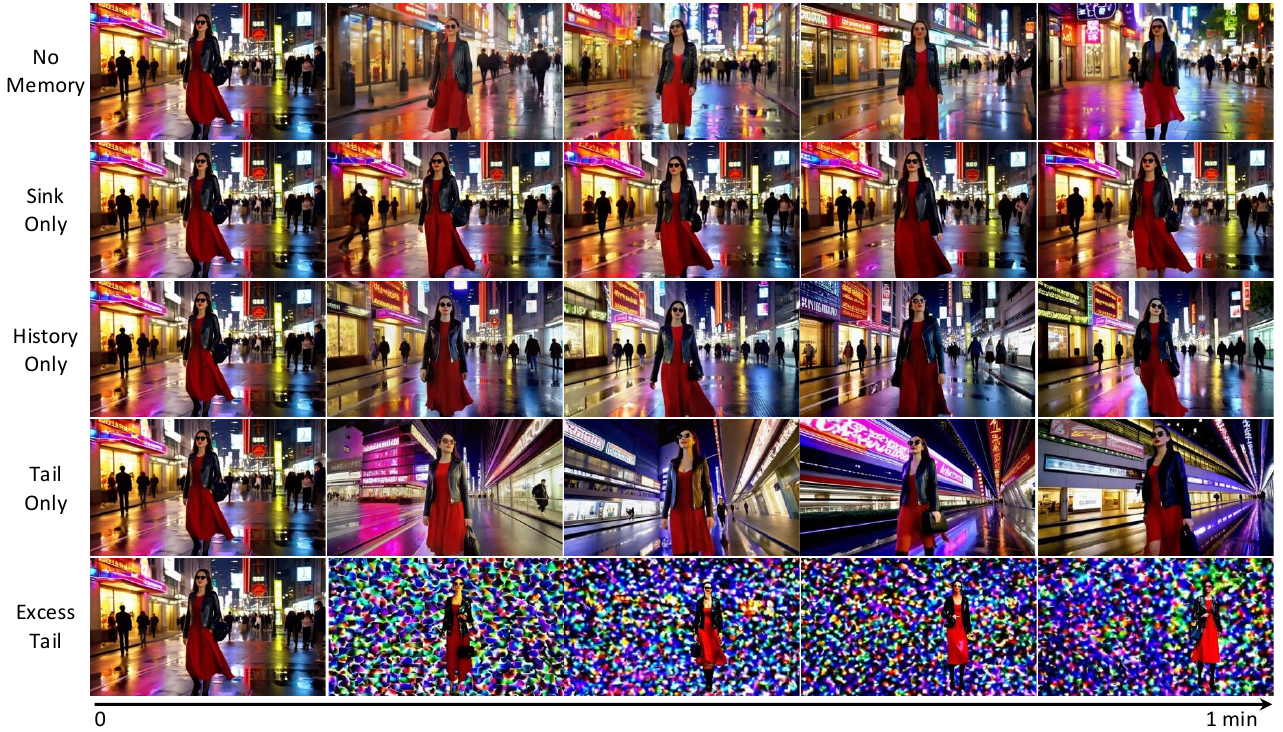}
    \vspace{-1em}
    \caption{Temporal memory roles during long-horizon rollout. Different memory configurations lead to distinct failure modes over time. Without memory, generation becomes fragmented. Sink-only conditioning preserves appearance but limits motion evolution. History-only conditioning enables motion variation but weakens identity consistency. Tail-only conditioning improves short-term continuity but destabilises long-term rollout, while excessive Tail leads to generation collapse. These results highlight the heterogeneous roles of temporal memory in balancing stability and motion dynamics.}
    \label{fig5}
    \vspace{-1.5em}
\end{figure}

\vspace{-1.0em}
\subsection{Temporal Memory Analysis for Long Video Extrapolation}
\vspace{-0.5em}

Despite the robustness gained from self-forcing training, long-horizon extrapolation still requires the model to repeatedly condition on previously synthesised frames. Under sliding-window inference, these frames are stored as KV memory and reused across rollout steps. Thus, long-video quality depends not only on the autoregressive model itself, but also on how temporal memory is organised. Existing AR diffusion methods commonly retain a dense chronological buffer within the attention window, implicitly treating historical frames as uniformly useful context. However, the empirical findings in Fig.~\ref{fig2} suggest that this assumption is overly restrictive: memory quantity and temporal placement affect motion evolution in different ways.

\begin{figure}[t]
    \vspace{-0.5em}
    \centering 
    \includegraphics[width=0.99\textwidth]{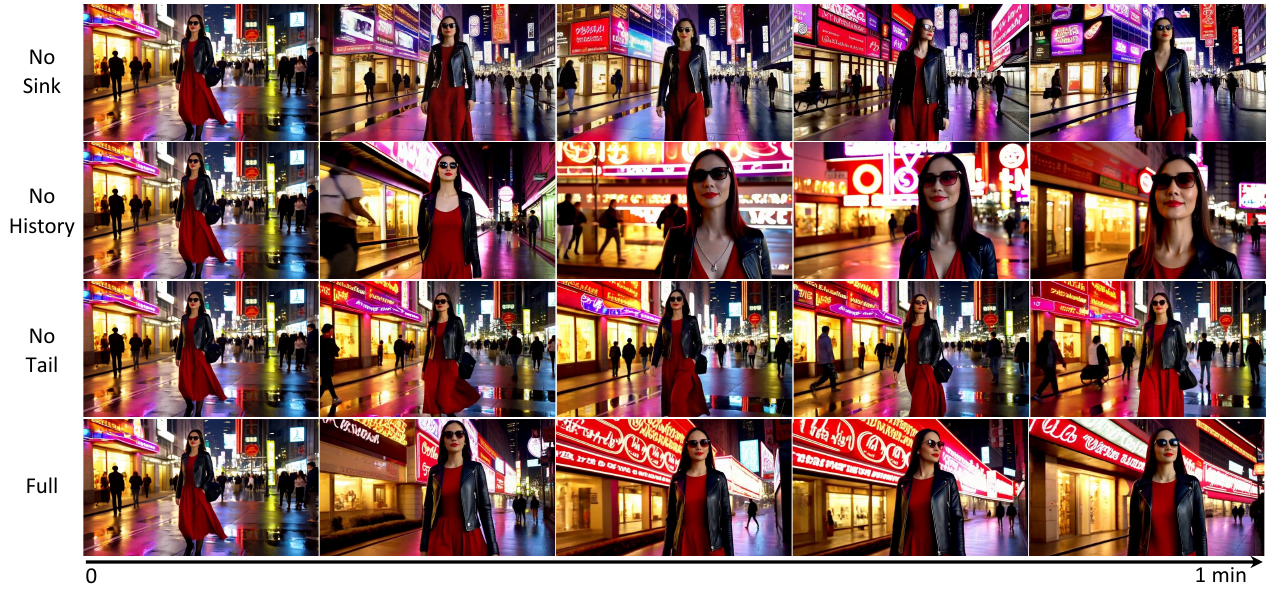}
    \vspace{-1em}
    \caption{Complementary roles of temporal memory components during long-horizon rollout. Removing individual components leads to distinct failure modes. Without Sink, the generation exhibits temporal drift. Without History, motion evolution becomes limited and repetitive. Without Tail, short-term continuity is weakened, resulting in rigid motion. A balanced combination of Sink, History, and Tail maintains both stability and dynamic evolution.}
    \label{fig6}
    \vspace{-1em}
\end{figure}

\noindent \textbf{Sensitivity to Memory Quantity and Temporal Placement.}
As shown in Fig.~\ref{fig2}(a–c), increasing the number of Sink, History, or Tail frames for conditioning does not monotonically improve generation quality. Visual fidelity remains relatively stable across memory sizes, whereas motion dynamics vary substantially and may degrade when additional frames are retained. This indicates that excessive memory can introduce redundant or overly constraining temporal cues rather than useful long-range guidance. Fig.~\ref{fig2}(d) further shows that temporal placement matters under a fixed memory budget: sampling history frames from different temporal positions produces noticeable changes in motion dynamics, while visual consistency and image quality remain largely unchanged. These results suggest that historical context is not interchangeable and that mid-range memory primarily affects structural motion rather than appearance stability.

\noindent \textbf{Functional Roles and Interdependence of Temporal Regions.}
To interpret these observations, we isolate different memory components during long-horizon rollout, as shown in Fig.~\ref{fig5}. Each component produces a distinct failure mode when used alone. Without memory, generation becomes fragmented over time. Sink-only conditioning preserves appearance but suppresses motion evolution, leading to repetitive dynamics. Tail-only conditioning improves local continuity but is insufficient for stable long-term rollout, and excessive Tail conditioning can cause collapse. History-only conditioning encourages motion variation but weakens identity and scene consistency.

We further examine the interdependence of these components by removing each one from a balanced memory configuration, as shown in Fig.~\ref{fig6}. Removing Sink leads to temporal drift, indicating the need for global anchors. Removing History limits motion evolution, confirming the importance of mid-range structure. Removing Tail weakens short-term continuity and produces rigid motion. In contrast, combining Sink, History, and Tail preserves both stability and dynamic evolution.
Taken together, these findings indicate that temporal memory is heterogeneous rather than a homogeneous chronological buffer. Effective long-horizon generation requires balancing complementary roles: \emph{early frames provide global anchors, recent frames maintain short-term continuity, and mid-range frames contribute structural motion cues}. This motivates a structured formulation of temporal memory, which we introduce next.

\vspace{-1.0em}
\subsection{Relaxed KV Memory}
\vspace{-0.5em}

\noindent \textbf{KV Memory Selection.}
Relaxed KV Memory decomposes temporal conditioning into three functional components:
a fixed Sink $\mathcal{S}$ serving as a long-term anchors,
a step-dependent Tail $\mathcal{T}_i$ ensuring short-term continuity at generation step $i$,
and a dynamically selected History set $\mathcal{H}_i$ drawn from an intermediate candidate region to stabilise and contextualise the current prediction.

At generation step $i$, the available historical frames are partitioned as an ordered sequence
\begin{equation}
\hat{x}^{<i} = [\mathcal{S}; \mathcal{M}_i^{cand}; \mathcal{T}_i],
\end{equation}
where $\mathcal{M}i^{cand}$ denotes the mid-range candidate frames, and $[\cdot;\cdot]$ denotes temporal concatenation.
Based on the empirical observation that informative history frames tend to appear in the later part of the mid-range region, as illustrated in Fig.~\ref{fig2}(d), we restrict candidate selection to the second half of $\mathcal{M}_i^{cand}$. Formally, we define a reduced candidate set
\begin{equation}
\tilde{\mathcal{M}}_i =
\left\{
h \in \mathcal{M}_i^{cand} \mid
\operatorname{idx}(h) \ge \tfrac{1}{2}|\mathcal{M}_i^{cand}|
\right\},
\end{equation}
where $\operatorname{idx}(h)$ denotes the temporal index of frame $h$ within $\mathcal{M}i^{cand}$.
For each candidate frame $h \in \tilde{\mathcal{M}}i$, we compute a representative key prototype
\begin{equation}
\tilde{K}h =
\operatorname{norm}
\left(
\frac{1}{|\Omega(h)|}
\sum{k \in \Omega(h)} K_k
\right),
\end{equation}
where $\Omega(h)$ denotes the set of tokens belonging to frame $h$. Similarly, aggregated Sink and Tail prototypes are defined as
$\tilde{K}{\mathcal{S}} = \operatorname{norm}\big(\mathcal{K}(\mathcal{S})\big)$ and $\tilde{K}{\mathcal{T}_i} = \operatorname{norm}\big(\mathcal{K}(\mathcal{T}_i)\big)$.

We then compute a stability score and a redundancy score:
$S(h) = \tilde{K}h^\top \tilde{K}{\mathcal{S}}$ and
$R(h) = \tilde{K}h^\top \tilde{K}{\mathcal{T}_i}$.
The relaxation score of candidate frame $h$ is defined as
\begin{equation}
r(h) = S(h) - \lambda R(h),
\end{equation}
where $\lambda$ controls the trade-off between global stability and local redundancy.

Finally, History frames are selected using Top-$K$ ranking
\begin{equation}
\mathcal{H}i =
\operatorname{TopK}
\big(
{r(h)}{h \in \tilde{\mathcal{M}}_i}
\big),
\end{equation}
and the memory used for conditioning is constructed as the ordered sequence
$\mathcal{M}_i = [\mathcal{S}; \mathcal{H}_i; \mathcal{T}_i]$.

\noindent \textbf{Extrapolation with Sparse Memory.}
In standard sliding-window inference, conditioning relies on a dense contiguous buffer of the most recent frames: $c^{<i} = \hat{x}^{i-L:i}$, where $L$ denotes the attention window size.

In contrast, Relaxed KV replaces this dense chronological buffer with a structured temporal memory
\begin{equation}
c^{<i} = \mathcal{M}_i = [\mathcal{S}; \mathcal{H}_i; \mathcal{T}_i],
\end{equation}
where the Sink $\mathcal{S}$ provides a fixed global anchor, while the History $\mathcal{H}_i$ and Tail $\mathcal{T}_i$ are dynamically updated during generation.
Specifically, at each rollout step, the model constructs $\mathcal{M}_i$ on the fly by selecting informative mid-range history frames according to the relaxation score $r(h)$, while always retaining the Sink anchors and the most recent Tail frames.
This design is motivated by our empirical observations in Fig.~\ref{fig2} and ~\ref{fig5}, which show that excess memory often introduces redundant context and constrains motion evolution. By retaining only a small set of informative history frames together with global anchors and recent context, Relaxed KV avoids repeatedly conditioning on redundant past frames.
The corresponding attention keys are therefore constructed as
\begin{equation}
\mathbf{K}{attn}^i = \mathcal{K}(\mathcal{M}_i).
\end{equation}
As a result, sparse memory conditioning mitigates error amplification during long-horizon rollout while preserving both global consistency and evolving motion dynamics.

\noindent \textbf{Position Embedding Adjustment.}
Relaxed KV Memory constructs a non-contiguous conditioning sequence
$\mathcal{M}_i = [\mathcal{S}; \mathcal{H}i; \mathcal{T}i]$,
which makes standard sliding-window RoPE resetting unsuitable.
We therefore adopt a hybrid positional indexing scheme that treats recent Tail and distant Sink or History memory differently.

For Tail, we preserve its actual temporal location by applying RoPE with absolute frame indices.
Let $i$ denote the current generation frame index and let $|\mathcal{T}i|$ be the number of Tail frames.
We define the Tail start index as $p{\mathcal{T}} = i - |\mathcal{T}i|$, and apply RoPE to Tail keys using indices ${p{\mathcal{T}},\ldots,i-1}$.
For Sink and History, instead of using their original absolute positions, we apply RoPE with relative indices anchored immediately before the Tail segment.
Let $|\mathcal{S}|$ and $|\mathcal{H}i|$ denote the number of Sink and History frames, and define $p{\mathcal{SH}} = p{\mathcal{T}} - (|\mathcal{S}| + |\mathcal{H}i|)$.
We then assign Sink plus History indices to the contiguous range ${p{\mathcal{SH}}, \ldots, p{\mathcal{T}}-1}$ and apply RoPE accordingly.

This adjustment enforces a consistent ordering of memory roles in positional space, with Sink and History preceding Tail, without forcing all selected memory frames to be locally contiguous within the sliding window. By keeping Tail on absolute positions while anchoring Sink and History relative to the current context, the model can leverage long-range anchors and mid-range structure without collapsing them into recent dynamics during window shifts.

\vspace{-1.5em}
\section{Experiments}
\vspace{-0.5em}
\subsection{Experimental Settings}
\vspace{-0.5em}
\noindent\textbf{Implementation Details.}
We implement chunk-wise Self Forcing~\cite{huang2025self} as our base model. Videos are generated autoregressively in chunks with a chunk size of 3 frames. For Relaxed KV Memory, we use 2 Sink frames, 1 Tail frame, and select 1 History frame from a candidate pool of 4 frames. The redundancy weight $\lambda$ is set to 2.0. All methods are evaluated under the same inference settings for fair comparison. Following previous work, all throughput experiments are conducted on a single NVIDIA H100 GPU.

\noindent \textbf{Evaluation.}
We evaluate long-horizon video generation using the VBench-Long benchmark~\cite{huang2025vbench++}, following the protocol adopted in recent autoregressive video diffusion works~\cite{cui2025self,yi2025deep}. Specifically, we use 128 prompts from MovieGen~\cite{polyak2024movie}, which are commonly used for long-video evaluation. Following prior work~\cite{huang2025self,yi2025deep,lu2025reward}, each prompt is refined using Qwen2.5-7B-Instruct~\cite{qwen2025qwen25technicalreport} to improve prompt clarity and diversity. We generate 30-second and 60-second videos and evaluate them with VBench-Long metrics, including subject consistency, background consistency, aesthetic quality, imaging quality, motion smoothness, and dynamic degree. In addition, we report CLIP-based temporal metrics to quantify drift and repetition across frames. Further implementation details are provided in the \textit{Appendix}.

\begin{table}[!t]
\begin{center}
\caption{Quantitative comparison of long-video generation on VBench-Long.}
\label{tab1}
\setlength{\tabcolsep}{2pt}
\resizebox{\columnwidth}{!}{%
\begin{tabular}{@{}c|c|c|cccccc|c@{}}
\toprule
\multirow{2}{*}{Methods} &
\multicolumn{1}{c|}{\multirow{2}{*}{\begin{tabular}[c]{@{}c@{}}Training\\ Free\end{tabular}}} &
\multirow{2}{*}{\begin{tabular}[c]{@{}c@{}}Throughput\\ (FPS)\end{tabular}} &
\multirow{2}{*}{\begin{tabular}[c]{@{}c@{}}Subject\\ Consistency\end{tabular}} &
\multirow{2}{*}{\begin{tabular}[c]{@{}c@{}}Background\\ Consistency\end{tabular}} &
\multirow{2}{*}{\begin{tabular}[c]{@{}c@{}}Aesthetic\\ Quality\end{tabular}} &
\multirow{2}{*}{\begin{tabular}[c]{@{}c@{}}Imaging\\ Quality\end{tabular}} &
\multirow{2}{*}{\begin{tabular}[c]{@{}c@{}}Motion\\ Smoothness\end{tabular}} &
\multirow{2}{*}{\begin{tabular}[c]{@{}c@{}}Dynamic\\ Degree\end{tabular}} &
\multirow{2}{*}{Average} \\
 & \multicolumn{1}{c|}{} &  &  &  &  &  &  &  &  \\ \midrule \midrule
\textbf{} & \multicolumn{9}{c}{30 seconds} \\ \midrule
CausVid \cite{yin2025slow} [CVPR'25]             
& $\times$ & 15.78 & 97.92 & 96.77 & 59.77 & 66.36 & 98.08 & 47.21 & 77.69 \\
Self Forcing \cite{huang2025self} [NeurIPS'25]   
& $\times$ & 15.78 & 97.34 & 96.47 & 59.44 & 68.58 & 98.63 & 36.62 & 76.18 \\
Rolling Forcing \cite{liu2025rolling} [ICLR'26] 
& $\times$ & 15.75 & \textbf{98.07} & \textbf{96.84} & \underline{60.75} & \textbf{70.73} & \underline{98.74} & 32.71 & 76.31 \\
LongLive \cite{yang2025longlive} [ICLR'26]       
& $\times$ & \textbf{18.16} & \underline{97.97} & \underline{96.83} & \textbf{61.51} & 69.07 & \textbf{98.76} & 45.55 & 78.28 \\
Deep Forcing \cite{yi2025deep} [ICML'26]          
& $\surd$ & 15.79 & 97.34 & 96.48 & 60.68 & \underline{69.31} & 98.27 & 57.56 & \underline{79.94} \\
\midrule
\textbf{CausVid + Relaxed KV Memory}                    
& $\surd$ & \underline{16.33} & 97.87 & 96.71 & 59.87 & 67.17 & 97.58 & \underline{59.88} & 79.85 \\
\textbf{Relax Forcing (Ours)}                    
& $\surd$ & \underline{16.33} & 96.99 & 96.12 & 60.13 & 68.50 & 97.80 & \textbf{65.67} & \textbf{80.87} \\
\midrule
\textbf{} & \multicolumn{9}{c}{60 seconds} \\ \midrule \midrule
CausVid \cite{yin2025slow} [CVPR'25]             
& $\times$ & 15.78 & 97.81 & 96.75 & 59.42 & 65.84 & 98.09 & 46.44 & 77.39 \\
Self Forcing \cite{huang2025self} [NeurIPS'25]   
& $\times$ & 15.78 & 96.31 & \textbf{96.82} & 56.45 & 66.33 & 98.21 & 31.98 & 74.35 \\
Rolling Forcing \cite{liu2025rolling} [ICLR'26]  
& $\times$ & 15.75 & \textbf{97.94} & \underline{96.76} & \underline{60.02} & \textbf{70.72} & \underline{98.71} & 32.50 & 76.11 \\
LongLive \cite{yang2025longlive} [ICLR'26]       
& $\times$ & \textbf{18.16} & \underline{97.85} & 96.74 & \textbf{61.29} & 69.11 & \textbf{98.75} & 43.49 & 77.87 \\
Deep Forcing \cite{yi2025deep} [ICML'26]           
& $\surd$ & 15.79 & 96.96 & 96.32 & 59.86 & \underline{69.27} & 98.23 & 57.19 & \underline{79.64} \\
\midrule
\textbf{CausVid + Relaxed KV Memory} 
& $\surd$ & \underline{16.33} & 97.70 & 96.62 & 59.27 & 66.49 & 97.53 & \underline{57.78} & 79.23 \\
\textbf{Relax Forcing (Ours)}                    
& $\surd$ & \underline{16.33} & 96.81 & 95.97 & 59.58 & 68.66 & 97.74 & \textbf{66.49} & \textbf{80.88} \\
\bottomrule
\end{tabular}
}
\vspace{-2.8em}
\end{center}
\end{table}

\vspace{-1.5em}
\subsection{Comparisons to State of the Art}
\vspace{-0.5em}
We compare Relax Forcing with recent autoregressive video diffusion methods on VBench-Long under both 30-second and 60-second settings. As shown in Tab.~\ref{tab1}, Relax Forcing achieves the highest overall score in both regimes. For 30-second videos, our method reaches 80.87\%, outperforming the strongest competing training-free method, Deep Forcing, by 0.93 percentage points. Under the more challenging 60-second setting, Relax Forcing achieves an overall score of 80.88\%, while other methods exhibit greater degradation, indicating that Relax Forcing is more robust during long-horizon rollout.

The largest improvement is observed in Dynamic Degree, where Relax Forcing achieves 65.67\% at 30 seconds and 66.49\% at 60 seconds. At the same time, subject consistency, background consistency, and imaging quality remain competitive with prior methods, indicating that stronger motion evolution is achieved without substantial degradation in visual fidelity. This supports our hypothesis that structured temporal memory can alleviate overly constrained dynamics caused by repeatedly conditioning on dense chronological history.

Importantly, the benefit of Relaxed KV Memory is not specific to Self Forcing. When directly applied to CausVid without retraining or architectural modification, it improves Dynamic Degree from 47.21\% to 59.88\% and the overall score from 77.69\% to 79.85\% at 30 seconds. At 60 seconds, Dynamic Degree increases from 46.44\% to 57.78\%, while the overall score improves from 77.39\% to 79.23\%. These consistent gains demonstrate that the proposed memory mechanism generalises to another autoregressive video diffusion baseline.

Finally, unlike Self Forcing~\cite{huang2025self}, Rolling Forcing~\cite{liu2025rolling}, and LongLive~\cite{yang2025longlive}, which improve long-horizon generation through training or architectural modifications, Relax Forcing operates entirely at inference time. Its structured sparse memory also reduces the effective attention length, providing additional efficiency benefits that we analyse in Sec.~\ref{sec:method} and in the latency ablation below.

We further evaluate challenging scenarios involving abrupt scene changes, multiple subjects, and high-speed or non-uniform motion, in which Relax Forcing achieves the highest overall score across both 30-second and 60-second settings; full results are provided in the Appendix.

\begin{table}[!t]
\begin{center}
\caption{Effect of different memory conditioning strategies.}
\label{tab2}
\setlength{\tabcolsep}{3pt}
\resizebox{0.97\columnwidth}{!}{%
\begin{tabular}{@{}c|cccccc|c@{}}
\toprule
Methods 
& \begin{tabular}[c]{@{}c@{}}Subject\\ Consistency\end{tabular}
& \begin{tabular}[c]{@{}c@{}}Background\\ Consistency\end{tabular}
& \begin{tabular}[c]{@{}c@{}}Aesthetic\\ Quality\end{tabular}
& \begin{tabular}[c]{@{}c@{}}Imaging\\ Quality\end{tabular}
& \begin{tabular}[c]{@{}c@{}}Motion\\ Smoothness\end{tabular}
& \begin{tabular}[c]{@{}c@{}}Dynamic\\ Degree\end{tabular}
& Average \\
\midrule
Baseline (Self Forcing) 
& 97.22 & 96.39 & 59.18 & 68.43 & \textbf{98.39} & 39.38 & 76.50 \\
+ Full Attention 
& \textbf{97.25} & \textbf{96.60} & 59.22 & 66.43 & 98.29 & 45.03 & 77.14 \\
+ Attention Sink 
& 97.18 & 96.40 & 59.80 & \textbf{68.58} & 98.32 & 54.37 & 79.11 \\
\textbf{+ Relaxed KV Memory (Ours)} 
& 96.99 & 96.12 & \textbf{60.13} & 68.50 & 97.80 & \textbf{65.67} & \textbf{80.87} \\
\bottomrule
\end{tabular}
}
\end{center}
\vspace{-1.5em}
\end{table}
\begin{table}[t]
\centering
\caption{Analysis of memory allocation. We independently vary the number of Sink, History, and Tail frames while keeping the other components fixed.}
\label{tabA}
\setlength{\tabcolsep}{4pt}
\resizebox{0.9\columnwidth}{!}{%
\begin{tabular}{c|c|ccccccc}
\toprule
Component & Size
& \makecell{Subject\\Consistency}
& \makecell{Background\\Consistency}
& \makecell{Aesthetic\\Quality}
& \makecell{Imaging\\Quality}
& \makecell{Motion\\Smoothness}
& \makecell{Dynamic\\Degree}
& Overall \\
\midrule
\multirow{4}{*}{Sink}
& 0 & 95.84 & 95.26 & 58.92 & 67.56 & 97.68 & 54.93 & 78.36 \\
& 1 & 96.85 & 95.98 & 59.29 & 68.66 & 97.80 & 63.54 & 80.36 \\
& 2 & 96.99 & 96.13 & 60.12 & 68.50 & 97.80 & 65.62 & \textbf{80.86} \\
& 3 & 97.46 & 96.46 & 60.87 & 69.12 & 98.13 & 58.55 & 80.10 \\
\midrule
\multirow{4}{*}{History}
& 0 & 96.98 & 96.10 & 60.33 & 68.53 & 97.81 & 61.15 & 80.15 \\
& 1 & 96.99 & 96.13 & 60.12 & 68.50 & 97.80 & 65.62 & \textbf{80.86} \\
& 2 & 97.12 & 96.21 & 60.20 & 68.80 & 97.95 & 63.89 & 80.69 \\
& 3 & 97.19 & 96.27 & 60.17 & 68.87 & 98.00 & 63.62 & 80.69 \\
\midrule
\multirow{4}{*}{Tail}
& 0 & 96.62 & 96.01 & 60.24 & 68.74 & 96.96 & 64.39 & 80.49 \\
& 1 & 96.99 & 96.13 & 60.12 & 68.50 & 97.80 & 65.62 & \textbf{80.86} \\
& 2 & 97.21 & 96.31 & 60.38 & 67.77 & 98.04 & 57.71 & 79.57 \\
& 3 & 97.17 & 96.36 & 60.26 & 66.64 & 98.03 & 55.86 & 79.05 \\
\bottomrule
\end{tabular}
}
\vspace{-1.5em}
\end{table}

\vspace{-1.0em}
\subsection{Ablation Analysis}
\vspace{-0.5em}
We conduct ablation studies on the 30-second setting to validate key design choices in Relax Forcing, including the structured memory design, History selection strategy, memory allocation, and Hybrid RoPE. Additional sensitivity analyses of the candidate pool size and redundancy weight $\lambda$ are provided in the Appendix.

\noindent \textbf{Effect of Structured Memory Design.}~We first analyse how different memory conditioning strategies influence long-horizon generation. Starting from the Self Forcing baseline, we compare full attention over historical frames, sink anchoring, and the proposed relaxed memory selection. As shown in Tab.~\ref{tab2}, enabling full attention slightly improves Dynamic Degree over the baseline, indicating that additional historical context can provide useful temporal cues. Introducing sink anchoring further improves Dynamic Degree, highlighting the importance of global anchors during long-horizon rollout. Finally, Relaxed KV Memory achieves the highest Dynamic Degree and overall score, while maintaining competitive visual consistency metrics. This demonstrates that selectively retaining role-aware historical context is more effective than relying on dense chronological memory alone. 

\begin{table}[!t]
\begin{center}
\caption{Comparison of different history selection strategies in Relaxed KV Memory.}
\label{tab3}
\setlength{\tabcolsep}{5pt}
\resizebox{0.93\columnwidth}{!}{%
\begin{tabular}{@{}c|ccc|ccc@{}}
\toprule
Methods 
& \begin{tabular}[c]{@{}c@{}}Imaging\\ Quality $\uparrow$ \end{tabular}
& \begin{tabular}[c]{@{}c@{}}Dynamic\\ Degree $\uparrow$ \end{tabular}
& \textbf{Overall} $\uparrow$
& \begin{tabular}[c]{@{}c@{}}Drift $\downarrow$ \end{tabular}
& \begin{tabular}[c]{@{}c@{}}Repetition $\downarrow$ \end{tabular}
& \begin{tabular}[c]{@{}c@{}}\textbf{Balance} $\downarrow$ \end{tabular} \\
\midrule
Baseline (Self Forcing) 
& 68.58 & 36.62 & 76.18 & 2.70 & \textbf{81.34} & 1.000 \\
Random Sampling 
& 68.55 & 62.47 & 80.42 & 2.15 & 83.59 & 0.805 \\
Fixed-Positional Sampling 
& 68.51 & 64.22 & 80.59 & 2.18 & 83.16 & 0.768 \\
Attention-Based Sampling 
& 68.80 & 63.18 & 80.54 & 2.17 & 83.20 & 0.765 \\
First-Half Sampling 
& \textbf{68.83} & 59.76 & 80.27 & \textbf{1.68} & 87.87 & 1.000 \\
\textbf{Relax Forcing (Ours)} 
& 68.50 & \textbf{65.67} & \textbf{80.87} & 2.13 & 83.33 & \textbf{0.745} \\
\bottomrule
\end{tabular}
}
\end{center}
\vspace{-2.5em}
\end{table}

\noindent \textbf{Effect of Memory Allocation.}~We further analyse memory allocation by independently varying the number of Sink, History, and Tail frames while keeping the remaining components fixed. As shown in Tab.~\ref{tabA}, the best overall performance is achieved with 2 Sink frames, 1 History frame, and 1 Tail frame. These results confirm that long-horizon generation benefits from a balanced memory composition rather than simply increasing the amount of retained context.

For \textit{Sink frames}, increasing the number from 0 to 2 improves the overall score from 78.36\% to 80.86\%, together with clear gains in Dynamic Degree and visual consistency. This indicates that Sink frames provide useful global anchors for preserving long-term identity and scene stability. However, using 3 Sink frames reduces Dynamic Degree from 65.62\% to 58.55\%, suggesting that excessive global anchoring may over-constrain motion evolution.
For \textit{History frames}, introducing 1 selected History frame improves Dynamic Degree from 61.15\% to 65.62\%, leading to the best overall performance. This supports the role of mid-range temporal context in motion progression. Increasing the number of selected History frames beyond 1 does not yield further gains; instead, larger History sets introduce redundant temporal cues and slightly reduce Dynamic Degree and the overall score.
For \textit{Tail frames}, using 1 Tail frame achieves the best balance between short-term continuity and motion dynamics. Increasing the number of Tail frames slightly improves some consistency-related metrics, but substantially degrades Dynamic Degree, dropping from 65.62\% to 55.86\% when the number of Tail frames increases from 1 to 3. This suggests that excessive reliance on recent frames can make rollout overly conservative and suppress long-term motion evolution.

\noindent \textbf{Effectiveness of History Selection Strategy.}~We evaluate several strategies for selecting History frames from the candidate memory pool, including random sampling, fixed-positional sampling, attention-based sampling, and selecting frames from the first half of the candidate region. As shown in Tab.~\ref{tab3}, all variants achieve similar imaging quality, but differ noticeably in terms of motion dynamics. Relax Forcing achieves the highest Dynamic Degree at 65.67\% and the best overall score at 80.87\%, outperforming the alternative selection strategies.

We also observe a trade-off between temporal drift and repetition. For example, First-Half Sampling achieves the lowest Drift but has high Repetition, while the Self Forcing baseline shows the opposite tendency. To quantify this trade-off, we introduce a normalised Balance score, defined as the sum of min-max-scaled Drift and Repetition. Relax Forcing achieves the lowest Balance score of 0.745, indicating the best observed trade-off among the evaluated strategies.  In particular, the attention-based variant provides a controlled comparison with attention-relevance-based selection: with the same Sink and Tail configuration, our relaxation criterion improves Dynamic Degree from 63.18\% to 65.67\% and reduces the Balance score from 0.765 to 0.745.

\noindent \textbf{Effect of Hybrid RoPE.}~Relaxed KV Memory introduces non-contiguous temporal context, making the original contiguous sliding-window positional treatment suboptimal. We therefore isolate the contribution of the proposed hybrid RoPE while keeping the same Relaxed KV memory structure.

As shown in Tab.~\ref{tab:hybrid_rope}, Relaxed KV Memory already provides a substantial improvement when used without the proposed Hybrid RoPE, increasing Dynamic Degree from 36.62\% to 60.06\% and the overall score from 76.18\% to 79.76\%. Introducing Hybrid RoPE further improves Dynamic Degree to 65.65\% and the overall score to 80.88\%. These results show that the primary gain comes from the role-aware sparse memory design, while Hybrid RoPE provides an additional benefit by better accommodating the non-contiguous temporal context introduced by Relaxed KV Memory.

\begin{table}[!t]
\begin{center}
\caption{Ablation results of Hybrid RoPE in Relax Forcing.}
\label{tab:hybrid_rope}
\setlength{\tabcolsep}{3pt}
\resizebox{0.97\columnwidth}{!}{%
\begin{tabular}{@{}c|cccccc|c@{}}
\toprule
Methods 
& \begin{tabular}[c]{@{}c@{}}Subject\\ Consistency\end{tabular}
& \begin{tabular}[c]{@{}c@{}}Background\\ Consistency\end{tabular}
& \begin{tabular}[c]{@{}c@{}}Aesthetic\\ Quality\end{tabular}
& \begin{tabular}[c]{@{}c@{}}Imaging\\ Quality\end{tabular}
& \begin{tabular}[c]{@{}c@{}}Motion\\ Smoothness\end{tabular}
& \begin{tabular}[c]{@{}c@{}}Dynamic\\ Degree\end{tabular}
& Average \\
\midrule
Self Forcing
& \textbf{97.34} & \textbf{96.47} & 59.44 & \textbf{68.58} & \textbf{98.63} & 36.62 & 76.18 \\
+ Relax Forcing w/o Hybrid RoPE
& 96.71 & 95.96 & \underline{59.92} & 67.84 & \underline{98.03} & \underline{60.06} & \underline{79.76} \\
+ \textbf{Relax Forcing w/ Hybrid RoPE}
& \underline{96.99} & \underline{96.13} & \textbf{60.18} & \underline{68.51} & 97.80 & \textbf{65.65} & \textbf{80.88} \\
\bottomrule
\end{tabular}
}
\end{center}
\vspace{-1.5em}
\end{table}
\begin{table}[t]
\centering
\small
\caption{Latency profiling comparison between the dense baseline and Relax Forcing. The baseline performs self-attention over 21 frames, while Relax Forcing reduces the effective attention length to 7 frames using structured KV memory.}
\label{tabC}
\setlength{\tabcolsep}{6pt}
\begin{tabular}{c|cc|c}
\toprule
Metric & Baseline (21f) & Relax Forcing (7f) & Speedup \\
\midrule
Flash Attention & 444.2 ms & 168.1 ms & 2.64$\times$ \\
KV Select + RoPE & 125.5 ms & 114.1 ms & 1.10$\times$ \\
Candidate Scoring & 0.0 ms & 3.4 ms & -- \\
KV Cache Management & 13.0 ms & 49.5 ms & 0.26$\times$ \\
Total Self-Attention & 678.5 ms & 430.8 ms & 1.58$\times$ \\
Block Generation Time & 848 ms & 627 ms & 1.35$\times$ \\
\midrule
Diffusion Time & 67654 ms & 49515 ms & 1.37$\times$ \\
End-to-End Time & 89122 ms & 70650 ms & 1.26$\times$ \\
\bottomrule
\end{tabular}
\vspace{-2em}
\end{table}

\noindent \textbf{Latency Analysis.}
We further analyse the inference efficiency of Relax Forcing by profiling its runtime against a dense-memory baseline under identical generation settings. All experiments generate one-minute videos on a single NVIDIA H100 GPU. The dense baseline performs self-attention over 18 historical frames together with the current 3-frame block, resulting in an effective attention length of 21 frames. In contrast, Relax Forcing uses a structured memory consisting of 2 Sink frames, 1 History frame, and 1 Tail frame. Together with the current block, this reduces the effective attention length to 7 frames.

As shown in Tab.~\ref{tabC}, Relax Forcing substantially reduces the cost of the dominant attention operations. In particular, FlashAttention latency decreases from 444.2\,ms to 168.1\,ms per block, corresponding to a 2.64$\times$ speedup. This reduction is consistent with the shorter effective attention length introduced by structured KV memory. Although Relax Forcing introduces an additional candidate-scoring step for selecting History frames, its overhead is only 3.4\,ms per block, which is small compared with the reduction in attention latency.

We also observe an increase in KV cache management time, from 13.0\,ms to 49.5\,ms, due to the additional candidate region maintained for history selection. Nevertheless, this overhead is outweighed by the reduction in attention computation. Overall, Relax Forcing reduces total self-attention time by 1.58$\times$, accelerates diffusion generation by 1.37$\times$, and achieves a 1.26$\times$ end-to-end speedup. These results show that structured KV memory improves long-horizon generation efficiency in addition to temporal quality.

Additional profiling for 30-second, 60-second, and 120-second generation shows that the FlashAttention speedup remains stable at approximately 2.5$\times$ as the rollout horizon increases; detailed results are provided in the Appendix.

\vspace{-1.0em}
\subsection{Human Evaluation}
\vspace{-0.5em}
\begin{wrapfigure}{r}{0.55\textwidth}
  \vspace{-2em}
  \centering
  \includegraphics[width=\linewidth]{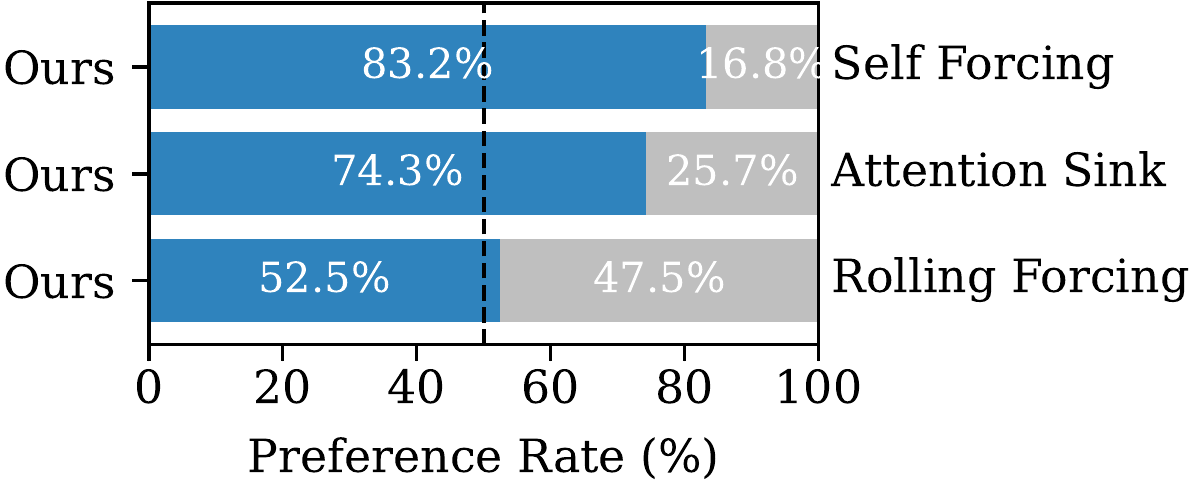}
  \vspace{-2em}
  \caption{Human preference study. Relax Forcing is preferred over competing methods in the user evaluation.}
  \label{fig:user_study_result}
  \vspace{-1em}
\end{wrapfigure}
To further assess perceptual quality, we conduct a pairwise user preference study comparing Relax Forcing with representative baselines. For each comparison, participants are shown, in random order, two videos generated from the same text prompt: one from Relax Forcing and one from a baseline method. Participants are asked to choose the preferred video according to overall perceptual quality, considering visual quality, motion quality, and text--video alignment. The results reported in Fig.~\ref{fig:user_study_result} are averaged over all evaluated prompts and participants. Additional implementation details are provided in the Appendix.

As shown in Fig.~\ref{fig:user_study_result}, Relax Forcing is strongly preferred over Self Forcing and Attention Sink, with preference rates of 83.2\% and 74.3\%, respectively. Compared with Rolling Forcing, Relax Forcing obtains a smaller but positive preference margin, with 52.5\% versus 47.5\%. These results suggest that the proposed structured memory mechanism improves perceived long-video quality, particularly over baselines that suffer from drift or constrained motion. The narrower margin against Rolling Forcing indicates that training-based rollout improvements remain competitive in human perception, while Relax Forcing provides complementary gains through training-free inference-time memory organisation.

\vspace{-1.5em}
\section{Conclusion}
\vspace{-0.5em}
In this work, we investigate the role of temporal memory in autoregressive video diffusion for long-horizon generation. Our analysis shows that increasing memory capacity alone is insufficient for reliable minute-scale synthesis; instead, the temporal organisation of historical context plays a critical role in balancing stability and motion evolution. Based on this insight, we introduce Relaxed KV Memory, the core component of Relax Forcing, which decomposes temporal context into Sink, Tail, and dynamically selected History components to enable structured sparse conditioning. Experiments on VBench-Long demonstrate that our training-free approach improves long-horizon motion dynamics and overall generation quality while reducing effective attention length and inference cost.

\vspace{-1.5em}
\section{Acknowledgments}
\vspace{-0.5em}
This research utilised Queen Mary’s Apocrita HPC facility, supported by QMUL Research-IT. Zengqun Zhao was funded by Queen Mary Principal’s PhD Studentships. This work was supported by Huawei Technologies Germany (Project 19655832: Uncertainty Aware Data Attribution Of Vision-Language Models).
\bibliography{reference}

\clearpage
\appendix
\counterwithin{figure}{section}
\counterwithin{table}{section}

\section*{Appendix}
This appendix provides supplementary material that supports the main paper. The contents are organised as follows:

\begin{itemize}[leftmargin=1.2em,itemsep=1pt,topsep=2pt,parsep=1pt]
\item \textbf{A: Algorithm Details.} Detailed Relaxed KV Memory selection and positional indexing.
\item \textbf{B: Experimental Settings.} Implementation details and evaluation protocol.
\item \textbf{C: Effect of History Frame Position.} Extended analysis of temporal History placement.
\item \textbf{D: Hyperparameter Sensitivity.} Candidate pool size and redundancy weight $\lambda$.
\item \textbf{E: Challenging-Prompt Robustness.} Evaluation on abrupt scene changes, multi-subject scenes, and high-speed/non-uniform motion.
\item \textbf{F: Extended Efficiency Analysis.} Runtime profiling for 30-second, 60-second, and 120-second generation.
\item \textbf{G: Human Evaluation Details.} User study protocol and annotation criteria.
\item \textbf{H: Qualitative Comparisons.} Additional visual comparisons and video examples.
\end{itemize}

\vspace{-1.5em}
\section{Algorithm Details}
\vspace{-0.5em}
\subsection{Relaxed KV Memory Selection}

Algorithm~\ref{alg:relaxed-kv} summarises the proposed Relaxed KV Memory selection procedure.
At each rollout step, the available historical frames are decomposed into three temporally ordered parts: a fixed Sink $\mathcal{S}$, a step-dependent Tail $\mathcal{T}_i$, and an intermediate candidate region $\mathcal{M}_i^{cand}$.
Based on the observation in Fig.~2(d) of the main paper, we restrict the selection of candidate History frames to the latter half of $\mathcal{M}_i^{cand}$ and rank candidate frames using the proposed relaxation score, which balances alignment with Sink against redundancy with Tail.

\begin{algorithm}[h]
\small
\DontPrintSemicolon
\caption{Relaxed KV Memory Selection}
\label{alg:relaxed-kv}

\KwIn{Generated frames $\hat{x}^{<i}$, Sink size $N_S$, Tail size $N_T$, History size $N_H$, redundancy weight $\lambda$}
\KwOut{Structured memory $\mathcal{M}_i$}

$\mathcal{S}\!\gets\!\hat{x}^{1:N_S}$;
$\mathcal{T}_i\!\gets\!\hat{x}^{i-N_T:i-1}$;
$\mathcal{M}_i^{\mathrm{cand}}\!\gets\!\hat{x}^{N_S+1:i-N_T-1}$;

$\tilde{\mathcal{M}}_i \!\gets\! {h \in \mathcal{M}_i^{\mathrm{cand}} \mid \mathrm{idx}(h) \ge \frac{1}{2}|\mathcal{M}_i^{\mathrm{cand}}|}$;

$\tilde{K}{\mathcal{S}} \!\gets\! \mathrm{norm}(\mathcal{K}(\mathcal{S}))$,
$\tilde{K}{\mathcal{T}_i} \!\gets\! \mathrm{norm}(\mathcal{K}(\mathcal{T}_i))$;

\ForEach{$h \in \tilde{\mathcal{M}}_i$}{
$\tilde{K}h \!\gets\! \mathrm{norm}\!\left(\frac{1}{|\Omega(h)|}\sum{k\in\Omega(h)} K_k\right)$;
$r(h) \!\gets\! \tilde{K}h^\top \tilde{K}{\mathcal{S}} - \lambda, \tilde{K}h^\top \tilde{K}{\mathcal{T}_i}$;
}

$\mathcal{H}i \!\gets\! \mathrm{TopK}{N_H}\big({r(h)}_{h\in\tilde{\mathcal{M}}_i}\big)$;

$\mathcal{M}_i \!\gets\! [\mathcal{S}; \mathcal{H}_i; \mathcal{T}_i]$;

\Return $\mathcal{M}_i$;
\end{algorithm}

\vspace{-1.5em}
\subsection{Sparse Extrapolation During Rollout}
\vspace{-0.5em}
In the standard sliding-window inference of Self Forcing \cite{huang2025self}, conditioning relies on a dense contiguous buffer of the most recent frames,
\begin{equation}
c^{<i}=\hat{x}^{i-L:i-1},
\end{equation}
where $L$ is the attention window size. 
In contrast, Relaxed KV replaces this dense memory with a structured memory
\begin{equation}
c^{<i}=\mathcal{M}_i=[\mathcal{S};\mathcal{H}_i;\mathcal{T}_i].
\end{equation}
At each rollout step, $\mathcal{H}_i$ is selected on the fly according to the relaxation score, while $\mathcal{S}$ is kept fixed and $\mathcal{T}_i$ is updated using the most recent frames. 
This design avoids repeatedly conditioning on redundant recent history and mitigates error amplification during long-horizon extrapolation.

\vspace{-1.0em}
\subsection{Hybrid Positional Indexing}
\vspace{-0.5em}

Because $\mathcal{M}_i$ is non-contiguous in time, standard sliding-window RoPE resetting~\cite{su2024roformer} is unsuitable.
We therefore apply RoPE differently to Tail and Sink/History memory:

\begin{itemize}[itemsep=1pt,topsep=2pt,parsep=1pt]
\item \textbf{Tail:} absolute frame indices are preserved so that the most recent context remains temporally grounded.
\item \textbf{Sink and History:} relative indices are assigned immediately before the Tail segment, preserving their role ordering without forcing them to be mapped according to their original absolute positions.
\end{itemize}

This hybrid indexing allows the model to exploit both global anchors and mid-range structure without collapsing them into short-term dynamics.

\vspace{-1.5em}
\section{Additional Experimental Settings}
\vspace{-0.5em}
\subsection{Inference Configuration}
\vspace{-0.5em}

Unless otherwise specified, all experiments use chunk-wise Self Forcing \cite{huang2025self} as the base model with a chunk size $U=3$ frames. 
For Relaxed KV Memory, we use:
\begin{itemize}[itemsep=1pt,topsep=2pt,parsep=1pt]
    \item Sink frames: 2
    \item Tail frames: 1
    \item History frames: 1
    \item Candidate pool size: 4
    \item Redundancy weight $\lambda$: 2.0
\end{itemize}

All methods are evaluated under the same inference settings for fair comparison.

\vspace{-1.0em}
\subsection{Prompt Set and Evaluation Protocol}

Following prior work on long-video autoregressive diffusion \cite{cui2025self,yi2025deep}, we evaluate on 128 prompts from MovieGen \cite{polyak2024movie}. 
To improve prompt clarity and diversity, the prompts are refined using Qwen2.5-7B-Instruct \cite{qwen2025qwen25technicalreport}. 
For robustness evaluation, we generate five videos for each prompt. 

We report the standard VBench-Long \cite{huang2025vbench++} metrics for both 30-second and 60-second generation. The evaluated metrics include:

\begin{itemize}[itemsep=1pt,topsep=2pt,parsep=1pt]
\item \textbf{Subject Consistency.}  
Measures whether the main subject maintains consistent appearance throughout the video. This metric is computed by measuring DINO feature similarity across frames.
\item \textbf{Background Consistency.}  
Evaluates the temporal consistency of background scenes by computing CLIP feature similarity across frames.
\item \textbf{Aesthetic Quality.}  
Assesses the visual aesthetics of generated frames using the LAION aesthetic predictor, which correlates with human perception of composition, color harmony, realism, and overall artistic quality.
\item \textbf{Imaging Quality.}  
Measures image-level distortions such as blur, noise, or exposure artifacts. This metric is computed using the MUSIQ image quality predictor trained on the SPAQ dataset.
\item \textbf{Motion Smoothness.}  
Evaluates the temporal smoothness of motion in generated videos using motion priors derived from a video frame interpolation model.
\item \textbf{Dynamic Degree.}  
Quantifies the magnitude of motion in generated videos. Optical flow estimated by RAFT is used to measure the degree of dynamic movement across frames.
\end{itemize}

For ablation experiments, we additionally report two complementary temporal metrics:

\begin{itemize}[itemsep=1pt,topsep=2pt,parsep=1pt]
\item \textbf{Drift.}  
Measures long-range semantic drift by computing the CLIP feature difference between the first and last 5-second clips of the generated video.
\item \textbf{Repetition.}  
Measures temporal repetition using the average CLIP2CLIP similarity among clips within the same generated video.
\item \textbf{Balance.}  
A normalised combination of Drift and Repetition that reflects the trade-off between temporal drift and excessive repetition. Lower values indicate a better balance.
\end{itemize}

\vspace{-1.5em}
\section{Effect of History Frame Position}
\vspace{-0.5em}
We further analyse how the temporal position of the selected History frame influences generation performance while fixing the memory configuration to Sink=2 and Tail=1, as shown in Tab.~\ref{tabB}. Overall, visual quality metrics such as Subject Consistency, Background Consistency, Aesthetic Quality, and Imaging Quality remain relatively stable across different history positions. In contrast, motion-related metrics show noticeable variation. Compared with the no-history setting, introducing a history frame generally improves Dynamic Degree, confirming the importance of mid-range temporal context for long-horizon extrapolation.

\begin{table}[t]
\centering
\small
\caption{Analysis on history frame selection positions while fixing the memory configuration to Sink=2 and Tail=1.}
\label{tabB}
\setlength{\tabcolsep}{4pt}
\resizebox{\columnwidth}{!}{%
\begin{tabular}{c|ccccccc}
\toprule
\makecell{History \\ Position}
& \makecell{Subject \\ Consistency}
& \makecell{Background \\ Consistency}
& \makecell{Aesthetic \\ Quality}
& \makecell{Imaging \\ Quality}
& \makecell{Motion \\ Smoothness}
& \makecell{Dynamic \\ Degree}
& Overall \\
\midrule
None
& 96.98 & 96.10 & 60.33 & 68.53 & 97.81 & 61.15 & 80.15 \\
\midrule
0
& 97.12 & 96.22 & 60.66 & 68.65 & 97.88 & 61.60 & 80.36 \\
1
& 96.99 & 96.13 & 60.41 & 68.58 & 97.80 & 63.34 & 80.54 \\
2
& 97.06 & 96.18 & 60.44 & 68.55 & 97.85 & 62.47 & 80.42 \\
3
& 97.07 & 96.19 & 60.46 & 68.61 & 97.85 & 62.00 & 80.36 \\
4
& 96.93 & 96.08 & 60.15 & 68.49 & 97.76 & 63.43 & 80.47 \\
5
& 96.99 & 96.12 & 60.14 & 68.48 & 97.79 & 63.54 & 80.51 \\
6
& 96.96 & 96.12 & 60.02 & 68.53 & 97.80 & 63.30 & 80.45 \\
7
& 96.93 & 96.09 & 60.02 & 68.61 & 97.76 & 63.62 & 80.51 \\
8
& 97.03 & 96.18 & 60.04 & 68.55 & 97.83 & 62.31 & 80.32 \\
\bottomrule
\end{tabular}
}
\vspace{-1.3em}
\end{table}

We further observe that positions in the middle-to-later portion of the candidate region tend to achieve stronger Dynamic Degree and better overall scores, whereas earlier positions provide weaker improvements. This observation supports our design choice of selecting informative mid-range history frames rather than relying on fixed chronological memory.

Overall, these results suggest that temporal memory should be carefully balanced rather than simply expanded. The best configuration in our experiments is obtained with Sink=2, History=1, and Tail=1, which achieves the highest overall score while maintaining strong motion dynamics.

\vspace{-1.5em}
\section{Hyperparameter Sensitivity}
\vspace{-0.5em}
\label{sec:app_hyperparameter}

We further study the sensitivity of Relax Forcing to two design parameters: the size of the candidate history pool and the redundancy weight $\lambda$. These experiments use the same 30-second evaluation protocol as the ablations in the main paper.

\vspace{-1.0em}
\subsection{Sensitivity to Candidate Pool Size}
\vspace{-0.5em}
\label{sec:app_candidate_pool}

We first study how the size of the candidate history pool affects long-horizon generation. Recall that Relaxed KV Memory selects one History frame from a candidate set sampled from the latter half of the intermediate temporal region. We vary the number of candidate frames while keeping the remaining memory configuration fixed to Sink$=2$, History$=1$, and Tail$=1$.

\begin{table}[!t]
\begin{center}
\caption{Effect of the size of candidate history frames used for Relaxed KV selection.}
\label{tab:app_candidate_pool}
\setlength{\tabcolsep}{5pt}
\resizebox{0.93\columnwidth}{!}{%
\begin{tabular}{@{}c|cccccc|c@{}}
\toprule
\begin{tabular}[c]{@{}c@{}}Candidates\\ Number\end{tabular}
& \begin{tabular}[c]{@{}c@{}}Subject\\ Consistency\end{tabular}
& \begin{tabular}[c]{@{}c@{}}Background\\ Consistency\end{tabular}
& \begin{tabular}[c]{@{}c@{}}Aesthetic\\ Quality\end{tabular}
& \begin{tabular}[c]{@{}c@{}}Imaging\\ Quality\end{tabular}
& \begin{tabular}[c]{@{}c@{}}Motion\\ Smoothness\end{tabular}
& \begin{tabular}[c]{@{}c@{}}Dynamic\\ Degree\end{tabular}
& Average \\
\midrule
2 & 96.98 & 96.12 & 60.16 & 68.52 & 97.79 & 65.55 & 80.85 \\
3 & \textbf{97.00} & 96.12 & \textbf{60.17} & \textbf{68.57} & \textbf{97.80} & 65.39 & 80.84 \\
4 & 96.99 & 96.13 & 60.12 & 68.50 & \textbf{97.80} & \textbf{65.62} & \textbf{80.86} \\
5 & \textbf{97.00} & \textbf{96.14} & 60.00 & 68.48 & 97.79 & 65.30 & 80.78 \\
7 & 96.94 & 96.09 & 60.14 & 68.46 & 97.77 & 63.44 & 80.47 \\
9 & 96.96 & 96.09 & 60.04 & 68.38 & 97.76 & 63.97 & 80.53 \\
\bottomrule
\end{tabular}
}
\vspace{-2em}
\end{center}
\end{table}
\begin{figure}[t]
\centering
\includegraphics[width=0.95\textwidth]{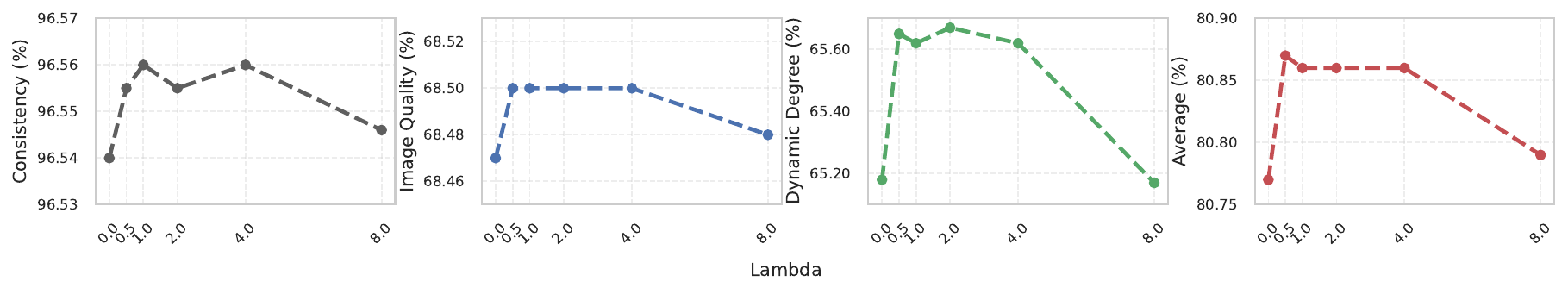}
\vspace{-1em}
\caption{Sensitivity to the redundancy weight $\lambda$ in Relaxed KV Memory. Performance remains stable across a moderate range of $\lambda$, while an excessively large redundancy penalty reduces motion dynamics and overall performance. We use $\lambda=2.0$ by default.}
\label{fig:app_lambda}
\vspace{-1em}
\end{figure}

As shown in Tab.~\ref{tab:app_candidate_pool}, performance is highly stable when the candidate pool contains 2–4 frames. The overall score varies by only 0.02 points across these settings, while Dynamic Degree remains above 65.3. This indicates that Relax Forcing does not rely on a precisely tuned candidate-pool size.

When the candidate pool becomes considerably larger, Dynamic Degree gradually decreases. For example, increasing the number of candidates from 4 to 7 reduces Dynamic Degree from 65.62 to 63.44. Since candidate frames are sampled from the latter half of the intermediate temporal region, a larger pool contains increasingly similar temporal content. Such redundant candidates provide less complementary motion information and may increase the likelihood of retaining correlated historical cues.

We therefore use four candidate frames by default, which provides a compact selection space while maintaining strong motion dynamics and overall generation quality.

\vspace{-1.0em}
\subsection{Robustness to Redundancy Weight $\lambda$}
\vspace{-0.5em}
\label{sec:app_lambda}

We next analyse the redundancy weight $\lambda$ in the relaxation score
\begin{equation}
r(h)=S(h)-\lambda R(h),
\end{equation}
which controls how strongly candidate History frames that resemble the recent Tail context are penalised.

As shown in Fig.~\ref{fig:app_lambda}, Relax Forcing remains stable across a moderate range of $\lambda$ values, with only minor variations in visual consistency, imaging quality, and the overall score. This indicates that the proposed selection criterion does not require precise tuning of the redundancy penalty.

When $\lambda$ becomes excessively large, however, Dynamic Degree decreases noticeably. A strong redundancy penalty can reject History frames that share useful short-term information with Tail, thereby removing temporal cues that are still important for coherent motion progression. Conversely, a moderate penalty reduces over-reliance on recent context while retaining sufficient temporal continuity.

We use $\lambda=2.0$ in all experiments, which provides the best observed balance between global anchoring, local continuity, and motion evolution.

\begin{table}[!t]
\begin{center}
\caption{Results on three challenging long-video generation settings.}
\label{tab:challenge_results}
\setlength{\tabcolsep}{2pt}
\resizebox{0.88\columnwidth}{!}{%
\begin{tabular}{@{}c|cccccc|c@{}}
\toprule
Method
& \begin{tabular}[c]{@{}c@{}}Subject\\ Consistency\end{tabular}
& \begin{tabular}[c]{@{}c@{}}Background\\ Consistency\end{tabular}
& \begin{tabular}[c]{@{}c@{}}Aesthetic\\ Quality\end{tabular}
& \begin{tabular}[c]{@{}c@{}}Imaging\\ Quality\end{tabular}
& \begin{tabular}[c]{@{}c@{}}Motion\\ Smoothness\end{tabular}
& \begin{tabular}[c]{@{}c@{}}Dynamic\\ Degree\end{tabular}
& Average \\
\midrule
\multicolumn{8}{c}{\textbf{Abrupt scene cuts / multi-shot editing, 30 seconds}} \\ \midrule
Self Forcing
& 95.11 & 95.35 & 59.70 & 67.12 & \underline{97.73} & 64.07 & 79.85 \\
Rolling Forcing
& \textbf{96.90} & \textbf{96.01} & \textbf{63.12} & \textbf{72.17} & \textbf{98.38} & 45.07 & 78.61 \\
Deep Forcing
& \underline{95.40} & \underline{95.47} & \underline{61.01} & 66.56 & 97.48 & \underline{67.73} & \underline{80.61} \\
Relax Forcing
& 95.32 & 95.21 & 60.92 & \underline{69.06} & 97.17 & \textbf{88.73} & \textbf{84.40} \\
\midrule
\multicolumn{8}{c}{\textbf{Abrupt scene cuts / multi-shot editing, 60 seconds}} \\ \midrule
Self Forcing
& 95.49 & \underline{95.59} & 58.73 & 66.28 & \underline{97.81} & 52.07 & 77.66 \\
Rolling Forcing
& \textbf{96.85} & \textbf{96.00} & \textbf{61.72} & \textbf{71.80} & \textbf{98.39} & 42.97 & 77.95 \\
Deep Forcing
& \underline{95.50} & 95.22 & 58.51 & 63.70 & 97.78 & \underline{59.33} & \underline{78.34} \\
Relax Forcing
& 95.32 & 95.15 & \underline{61.08} & \underline{68.67} & 97.06 & \textbf{86.47} & \textbf{83.96} \\
\midrule
\multicolumn{8}{c}{\textbf{Multi-subject generation, 30 seconds}} \\ \midrule
Self Forcing
& \underline{97.17} & 96.25 & 59.15 & \underline{72.38} & \underline{98.16} & 59.40 & 80.42 \\
Rolling Forcing
& \textbf{97.94} & \textbf{96.80} & \textbf{60.91} & \textbf{74.29} & \textbf{98.43} & 56.53 & 80.81 \\
Deep Forcing
& 96.80 & \underline{96.33} & 60.02 & 71.18 & 97.64 & \underline{70.20} & \underline{82.03} \\
Relax Forcing
& 96.47 & 95.96 & \underline{60.87} & 70.88 & 96.99 & \textbf{77.33} & \textbf{83.08} \\
\midrule
\multicolumn{8}{c}{\textbf{Multi-subject generation, 60 seconds}} \\ \midrule
Self Forcing
& \underline{96.90} & 96.04 & 55.38 & 69.31 & \underline{98.31} & 55.57 & 78.59 \\
Rolling Forcing
& \textbf{97.87} & \textbf{96.70} & \textbf{61.11} & \textbf{74.07} & \textbf{98.38} & 60.37 & \underline{81.42} \\
Deep Forcing
& 96.66 & \underline{96.07} & 59.04 & \underline{71.33} & 97.46 & \underline{65.20} & 80.96 \\
Relax Forcing
& 96.49 & 96.00 & \underline{60.78} & 71.12 & 96.97 & \textbf{77.87} & \textbf{83.20} \\
\midrule
\multicolumn{8}{c}{\textbf{High-speed / non-uniform motion, 30 seconds}} \\ \midrule
Self Forcing
& \underline{95.74} & \underline{95.31} & 56.59 & 62.18 & \underline{97.38} & 68.73 & 79.32 \\
Rolling Forcing
& \textbf{96.97} & \textbf{95.97} & \textbf{59.64} & \textbf{68.38} & \textbf{97.73} & 62.20 & 80.15 \\
Deep Forcing
& 95.64 & 95.29 & 57.18 & 61.16 & 96.96 & \underline{76.87} & \underline{80.52} \\
Relax Forcing
& 95.31 & 95.06 & \underline{57.58} & \underline{62.32} & 96.25 & \textbf{92.27} & \textbf{83.13} \\
\midrule
\multicolumn{8}{c}{\textbf{High-speed / non-uniform motion, 60 seconds}} \\ \midrule
Self Forcing
& 95.53 & \underline{95.33} & 54.15 & 60.84 & \underline{97.68} & 57.90 & 76.91 \\
Rolling Forcing
& \textbf{96.78} & \textbf{95.92} & \textbf{59.21} & \textbf{68.48} & \textbf{97.85} & 59.67 & \underline{79.65} \\
Deep Forcing
& \underline{95.77} & 95.26 & 56.49 & 59.86 & 97.21 & \underline{67.83} & 78.74 \\
Relax Forcing
& 95.18 & 95.08 & \underline{57.53} & \underline{62.59} & 96.24 & \textbf{90.83} & \textbf{82.91} \\
\bottomrule
\end{tabular}
}
\vspace{-3em}
\end{center}
\end{table}

\vspace{-1.5em}
\section{Challenging-Prompt Robustness}
\vspace{-0.5em}
\label{sec:app_challenging}
To evaluate robustness beyond the standard prompt distribution, we construct three challenging subsets covering:
(i) abrupt scene cuts and multi-shot editing,
(ii) multi-subject generation, and
(iii) high-speed or non-uniform motion.
For each setting, we select 30 prompts from MovieGen and evaluate both 30-second and 60-second generation using the same VBench-Long metrics.

As shown in Tab.~\ref{tab:challenge_results}, Relax Forcing achieves the highest overall score across all six challenging settings.
For abrupt scene changes, it achieves Average scores of 84.40 and 83.96 at 30s and 60s, respectively.
For multi-subject generation, the corresponding scores are 83.08 and 83.20, while for high-speed/non-uniform motion they are 83.13 and 82.91.
The largest gains are again observed in Dynamic Degree.
Relax Forcing reaches 88.73/86.47 for abrupt scene changes, 77.33/77.87 for multi-subject generation, and 92.27/90.83 for high-speed motion at 30s/60s.
These results indicate that role-aware memory remains effective under challenging long-horizon dynamics and is not limited to standard prompt configurations.

\vspace{-1.5em}
\section{Extended Efficiency Analysis}
\vspace{-0.5em}
\label{sec:app_efficiency}

The main paper reports detailed latency profiling for one-minute generation.
We additionally evaluate whether the attention-efficiency benefit of Relaxed KV Memory remains stable as the rollout horizon increases.

\begin{table}[!t]
\begin{center}
\caption{FlashAttention runtime comparison across different generation lengths.}
\label{tab:app_efficiency}
\setlength{\tabcolsep}{4pt}
\resizebox{0.8\columnwidth}{!}{%
\begin{tabular}{c|c|ccc|c}
\toprule
Length & Configuration
& \begin{tabular}[c]{@{}c@{}}FlashAttention\\ (ms)\end{tabular}
& \begin{tabular}[c]{@{}c@{}}\% of\\ Diffusion\end{tabular}
& \begin{tabular}[c]{@{}c@{}}Diffusion\\ Time (ms)\end{tabular}
& \begin{tabular}[c]{@{}c@{}}Attention\\ Speedup\end{tabular} \\
\midrule
30s & Relaxed KV & 7,273  & 29.8\% & 24,467  & -- \\
30s & Dense-21   & 17,848 & 53.5\% & 33,358  & 2.45$\times$ \\
\midrule
60s & Relaxed KV & 14,689 & 29.0\% & 50,781  & -- \\
60s & Dense-21   & 37,185 & 54.2\% & 68,588  & 2.53$\times$ \\
\midrule
120s & Relaxed KV & 30,000 & 27.9\% & 107,738 & -- \\
120s & Dense-21   & 75,612 & 54.6\% & 138,516 & 2.52$\times$ \\
\bottomrule
\end{tabular}
}
\vspace{-3em}
\end{center}
\end{table}

As shown in Tab.~\ref{tab:app_efficiency}, the FlashAttention speedup remains stable as generation length increases: 2.45$\times$ at 30 seconds, 2.53$\times$ at 60 seconds, and 2.52$\times$ at 120 seconds.
Moreover, FlashAttention accounts for only 27.9--29.8\% of diffusion time with Relaxed KV Memory, compared with approximately 53.5--54.6\% for the dense 21-frame baseline.
These results show that the computational benefit of structured sparse memory persists over longer autoregressive rollouts rather than being specific to a single video duration.

\vspace{-1.5em}
\section{Human Evaluation Details}
\vspace{-0.5em}

We conduct a user preference study to further assess perceptual quality.
Participants are shown generated videos from Relax Forcing and comparison methods and are asked to select the preferred result under the following criteria:

\begin{itemize}[itemsep=1pt,topsep=2pt,parsep=1pt]
\item \textbf{Visual Quality (VQ):} sharpness, realism, and overall visual plausibility.
\item \textbf{Motion Quality (MQ):} temporal smoothness, natural motion evolution, and absence of drift or repetition.
\item \textbf{Text–Video Alignment (TA):} consistency between the generated video and the text prompt.
\end{itemize}

For each prompt, videos from different methods are presented in randomised order to avoid position bias.
Each comparison is independently annotated by multiple participants, and the final preference score is computed as the percentage of pairwise wins.
We compare Relax Forcing against Self Forcing, Attention Sink, and Rolling Forcing, as reported in the main paper.

\vspace{-1.5em}
\section{Additional Qualitative Comparisons}
\vspace{-0.5em}
We provide additional qualitative comparisons with representative methods, including Self Forcing, Attention Sink, Rolling Forcing, and our Relax Forcing. These qualitative results highlight three recurring patterns:

\begin{itemize}[itemsep=1pt,topsep=2pt,parsep=1pt]
\item \textbf{Temporal Drift.} Methods with insufficient long-range anchoring often exhibit gradual semantic drift, including changes in identity, scene layout, or object appearance over time.
\item \textbf{Over-Constrained Motion.} Methods that rely heavily on recent context or dense chronological memory may maintain a stable appearance but produce repetitive or rigid motion.
\item \textbf{Balanced Evolution.} Relax Forcing preserves global consistency while allowing sustained motion evolution, producing videos that remain coherent without collapsing into either drift or repetition.
\end{itemize}

We also include additional video demos in the supplementary material to illustrate these temporal differences more clearly than static frames alone.

\begin{figure}[t]
    \centering 
    \includegraphics[width=0.95\textwidth]{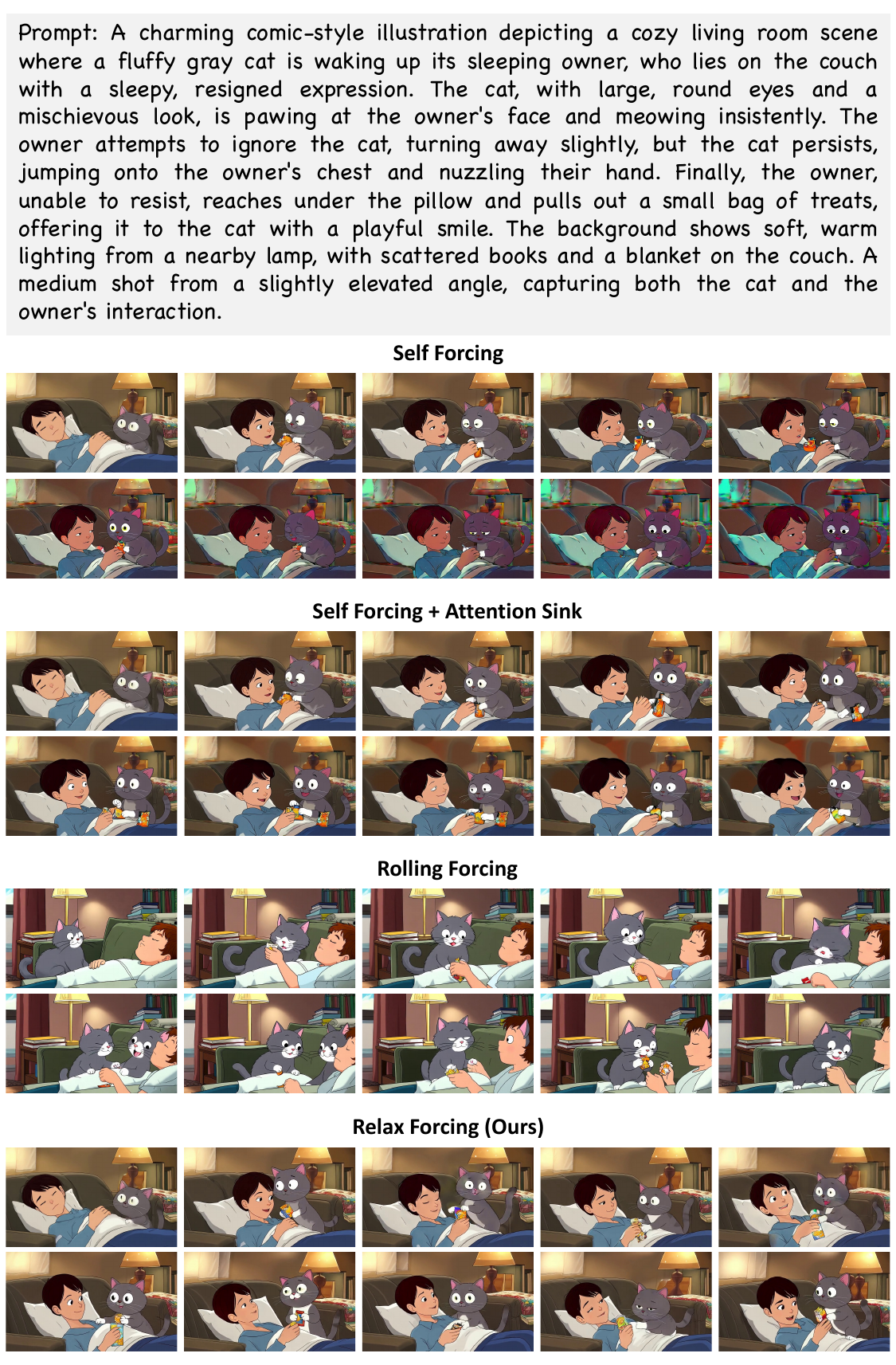}
    \vspace{-1em}
    \caption{Qualitative comparison of long-video generation under different methods.}
    \label{figB}
\end{figure}
\begin{figure}[t]
    \centering 
    \includegraphics[width=0.95\textwidth]{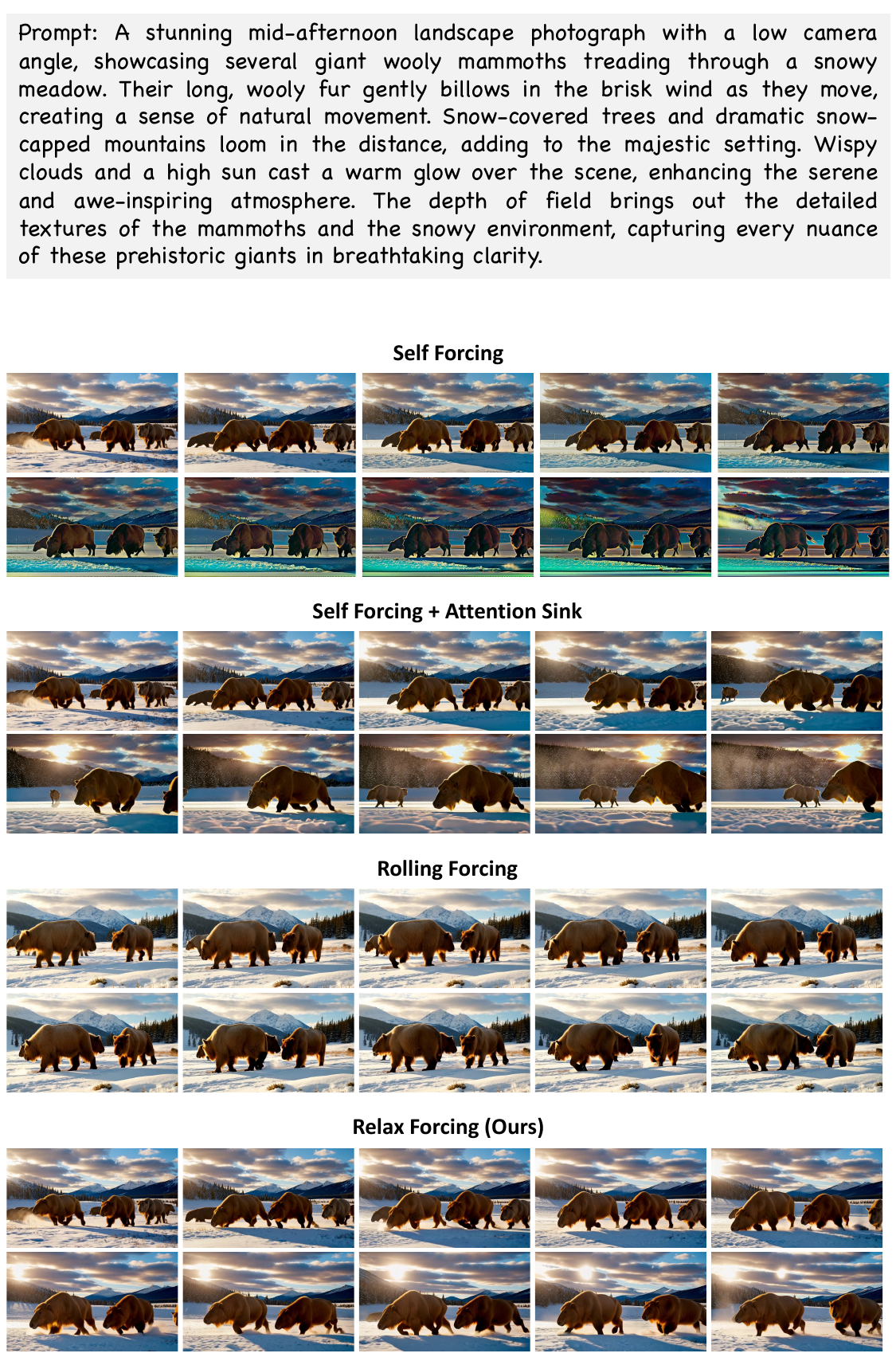}
    \vspace{-1em}
    \caption{Qualitative comparison of long-video generation under different methods.}
    \label{figB}
\end{figure}
\begin{figure}[t]
    \centering 
    \includegraphics[width=0.95\textwidth]{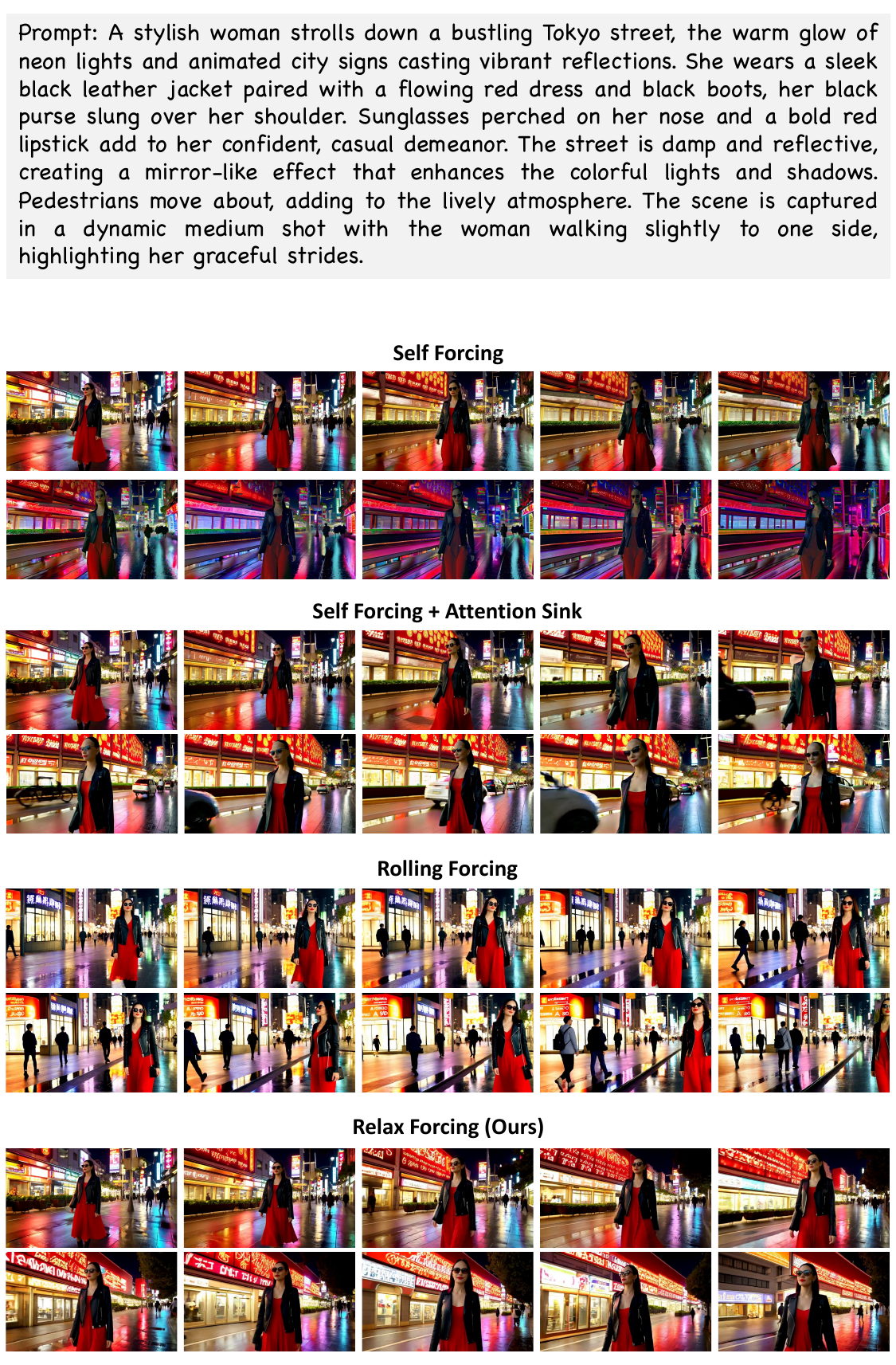}
    \vspace{-1em}
    \caption{Qualitative comparison of long-video generation under different methods.}
    \label{figB}
\end{figure}
\begin{figure}[t]
    \centering 
    \includegraphics[width=0.95\textwidth]{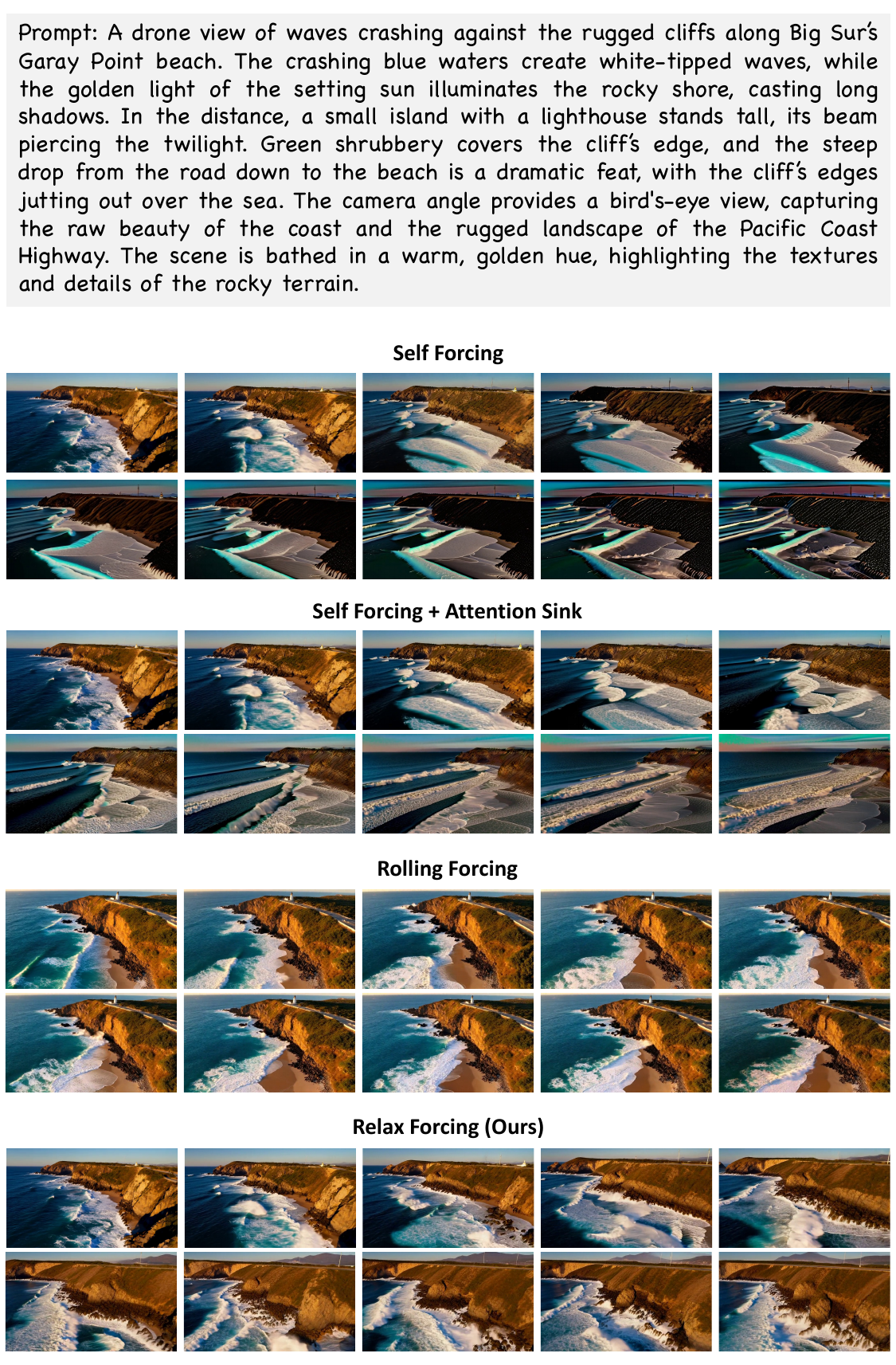}
    \vspace{-1em}
    \caption{Qualitative comparison of long-video generation under different methods.}
    \label{figB}
\end{figure}
\begin{figure}[t]
    \centering 
    \includegraphics[width=0.95\textwidth]{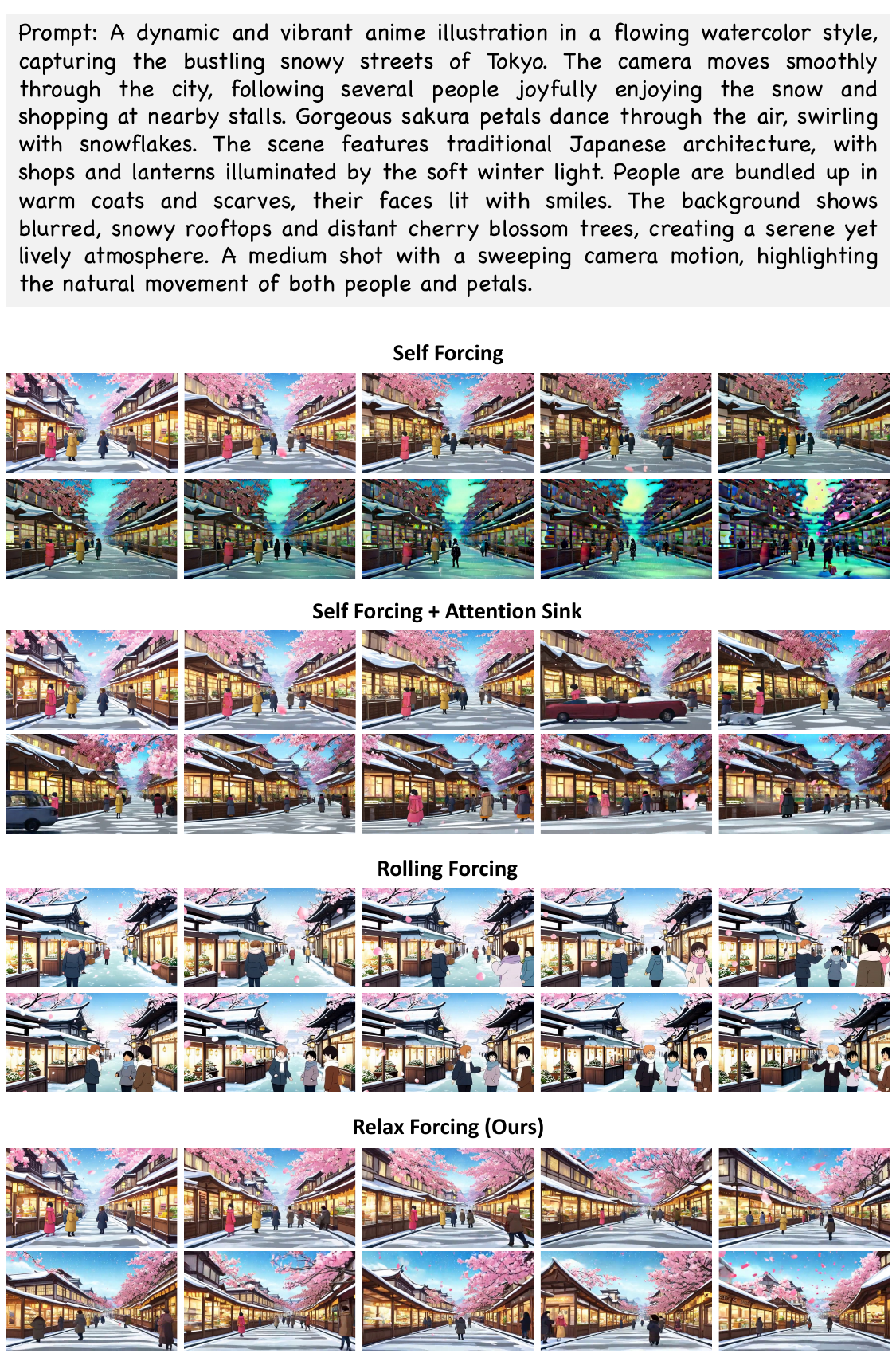}
    \vspace{-1em}
    \caption{Qualitative comparison of long-video generation under different methods.}
    \label{figB}
\end{figure}

\end{document}